\documentclass[lettersize,journal]{IEEEtran}
\usepackage{amsmath,amsfonts}
\usepackage{amsthm}
\usepackage{algorithmic}
\usepackage{algorithm}
\usepackage{bm}
\usepackage{array}
\usepackage{amssymb}
\usepackage{makecell}
\usepackage{color}
\usepackage{amsmath}
\usepackage[caption=false,font=normalsize,labelfont=sf,textfont=sf]{subfig}
\usepackage{textcomp}
\usepackage{stfloats}
\usepackage{url}
\usepackage{verbatim}
\usepackage{graphicx}
\usepackage{cite}
\newtheorem{corollary}{Corollary}
\newtheorem{thm}{Theorem}
\usepackage{url}

\begin{document}

\title{Class-Level Logit Perturbation}

\author{
         Mengyang Li*, Fengguang Su*, Ou Wu, Ji Zhang
\thanks{Manuscript received August, 2022. * denotes the equal contribution. Corresponding author: Ou Wu. \\
M. Li, F. Su and O. Wu are with National Center for Applied Mathematics, Tianjin University, Tianjin 300072, China (e-mail: limengyang@tju.edu.cn; fengguangsu@tju.edu.cn; wuou@tju.edu.cn). \\
J. Zhang is with University of Southern Queensland, Queensland 4350, Australia (e-mail: Ji.Zhang@usq.edu.au).}
}

\markboth{}%
{Shell \MakeLowercase{\textit{et al.}}: Class-Level Logit Perturbation}

\renewcommand\thethm{\arabic{thm}}
\setcounter{thm}{0}
\maketitle

\begin{abstract}
Features, logits, and labels are the three primary data when a sample passes through a deep neural network. Feature perturbation and label perturbation receive increasing attention in recent years. They have been proven to be useful in various deep learning approaches. For example, (adversarial) feature perturbation can improve the robustness or even generalization capability of learned models. However, limited studies have explicitly explored for the perturbation of logit vectors. This work discusses several existing methods related to  class-level logit perturbation. A unified viewpoint between positive/negative data augmentation and loss variations incurred by logit perturbation is established. A theoretical analysis is provided to illuminate why class-level logit perturbation is useful. Accordingly, new methodologies are proposed to explicitly learn to perturb logits for both single-label and multi-label classification tasks. Extensive experiments on benchmark image classification data sets and their long-tail versions indicated the competitive performance of our learning method. As it only perturbs on logit, it can be used as a plug-in to fuse with any existing classification algorithms. All the codes are available at https://github.com/limengyang1992/lpl.
\end{abstract}


\section{Introduction}
\IEEEPARstart{T}{here} are several main paradigms (which may overlap) among numerous deep learning studies, including new network architecture, new training loss, new training data perturbation scheme, and new learning strategy (e.g., weighting). Training data perturbation mainly refers to feature and label perturbations.

In feature perturbation, many data augmentation tricks can be viewed as feature perturbation methods when the input is the raw feature (i.e., raw samples). For example, cropped or rotated images can be seen as the perturbed samples of the raw images in computer vision; sentences with modified words can also be seen as the perturbed texts in text classification. Another well-known feature perturbation technique is about the generation of adversarial training samples~\cite{madry2017towards}, which attracts great attention in various AI applications especially in computer vision~\cite{xie2020adversarial} and natural language processing~\cite{jin2020bert}. Adversarial samples are those that can fool the learned models. They can be obtained by solving the following objective function:
\begin{equation}
\bm{x}_{adv} = \bm x + \arg\mathop {\max }\limits_{\left\| \bm \delta  \right\| \le \epsilon } l(f(\bm{x} + \bm \delta ),\bm y),\label{commomadv}
\end{equation}
where $\bm x$ is the input or the hidden feature; $\bm \delta$ is the perturbation term; $\epsilon$ is the perturbation bound; $\bm y$ is the one-hot label; and $\bm x_{adv}$ is the generated adversarial sample. A number of methods have been proposed to optimize Eq. (\ref{commomadv})~\cite{goodfellow2014explaining, madry2017towards}. Adversarial samples can be used to train more robust models.

In label perturbation, the labels are modified or corrected to avoid overfitting and noises. For example, a popular yet simple training trick, label smoothing~\cite{szegedy2016rethinking}, generates a new label for each sample according to $\bm y' = \bm y + \lambda (\frac{\bm I}{C} - \bm y)$, where $\bm y$ is the one-hot vector label; $C$ is the number of categories; $\bm I$ is a vector with all elements equaling to 1; $(\frac{\bm I}{C} - \bm y)$ is the perturbation term; and $\lambda$ is a hyper-parameter. Other methods such as Boostrapping loss~\cite{reed2014training}, label correction~\cite{patrini2017making,wang2021proselflc}, and Meta label corrector~\cite{wu2020learning} can be seen as a type of label perturbation. Mixup~\cite{zhang2017mixup} can be attributed to the combination of feature and label perturbation.

Logit vectors (or logits) are the outputs of the final feature encoding layer in almost all deep neural networks (DNNs). Although logits are important in the DNN data pipeline, only several learning methods in data augmentation and long-tail classification directly (without optimization) or implicitly employ class-level logit perturbation. Based on the loss analysis of these methods, the loss variations incurred by logit perturbation are highly related to the purpose of positive/negative augmentation\footnote{In this study, negative augmentation denotes the augmentation which aims to reduce the (relative) performances of some categories. Accordingly, existing augmentation methods are positive.} on training data. A theoretical analysis is conducted to reveal the connections among loss variations, performance improvements, and class-level logit perturbation. Accordingly, new methodologies are proposed to learn a class-level logit perturbation (LPL) for single-label and multi-label learning tasks, respectively, in this study. Extensive experiments are run on benchmark data sets for single-label classification and multi-label classification tasks. The results show the competitiveness of our methodologies.


Parts of the results in this paper were published originally in its conference version \cite{li2022logit}.  In our conference version, several classical methods are rediscussed in terms of logit perturbation and positive/negative augmentation. A new method is proposed to learn to perturb logits which can be used in implicit data augmentation and long-tail classification contexts for single-label classification tasks. Experimental results show that our method outperforms existing state-of-the-art methods related to logit perturbation in both contexts. This paper extends our earlier work in several important aspects:

\begin{itemize}

\item We conduct a theoretical analysis for the roles of logit perturbation-based explicit negative and positive augmentations in learning for binary classification tasks. Two typical scenarios, namely,  class imbalance and variance imbalance, are considered in our analysis.

\item We extend our LPL algorithm to the multi-label classification, which contains both class and variance imbalances, and empirically validate its effectiveness on multi-label classification benchmarks.

\item   Extensive experiments on large-scale long-tail data sets such as iNaturalist are performed. Our method LPL still achieves competitive results.


\end{itemize}

\section{Related Work}
\subsection{Data Augmentation}
Data augmentation is prevailed in almost all deep learning approaches. In its early stage, heuristic operations on raw samples are utilized, such as image flip, image rotation, and word replacing in sentences. Recently, advanced tricks are investigated, such as mixup~\cite{zhang2017mixup}, semantic data augmentation~\cite{wang2019implicit}, and meta semantic augmentation\cite{li2021metasaug}. In mixup, given a sample $\{\bm x_1, \bm y_1\}$, its perturbation term is $\left\{ {\lambda \left( {\bm x_2 - \bm x_1} \right), \lambda \left( {\bm y_2 - \bm y_1} \right)} \right\}$, where $\lambda$ is a random parameter (not a hyper-parameter), and $\{\bm x_2, \bm y_2\}$ is another randomly selected sample. Hu et al.~\cite{hu2019learning} introduce reinforcement learning to automatically augment data.

In this study, existing data augmentation is called positive data augmentation. Negative data augmentation, which is proposed in this study, may be helpful when we aim to restrain the (relative) performance of certain categories (e.g., to keep fairness in some tasks).

\subsection{Long-tail Classification}
Real data usually conform to a skewed or even a long-tail distribution. In long-tail classification, the proportions of tail samples are considerably small compared with those of head samples. Long-tail classification may be divided into two main strategies. The first strategy is to design new network architectures. Zhou et al.~\cite{zhou2020bbn} design a bilateral-branch network to learn the representations of head and tail samples. The second strategy is to modify the training loss. In this way, the weighting scheme~\cite{fan2017learning} is the most common practice. Relatively larger weights are exerted on the losses of the tail samples. Besides weighting, some recent studies modify the logits to change the whole loss, such as logit adjustment (LA)~\cite{menon2020long}. This new path achieves higher accuracies in benchmark data corpora compared with weighting~\cite{wu2021adversarial}.

\subsection{Multi-label Classification}
Real data usually also contain multiple objectives. Unlike the two single-label classification tasks mentioned above, there are two main challenges in multi-label classification tasks, namely, the co-occurrence of labels and the dominance of negative samples \cite{wu2020dist,guo2021long}. Li et al.\cite{8766125}  introduce a novel and effective deep metric learning method, which explores the relationship of images and labels by learning a two-way deep distance metric over two embedding spaces. Wei et al.\cite{8830456} investigate the impact of labels on evaluation metrics for large-scale multi-label learning and propose to restrain labels that have less impact on performance to speed up prediction and reduce model complexity. Wu et al.\cite{wu2020dist} perturb logits to emphasize the positive samples of tail categories to prevent class-specific overfitting. In multi-label classification task, weighting scheme \cite{lin2017focal} is also a typically used method. 

\subsection{Adversarial Training}
Adversarial training is an important way to enhance the robustness of neural networks \cite{deng2021adversarial,cui2021learnable}. The most important step in adversarial training is to generate adversarial training examples in Eq. (\ref{commomadv}), which can be used to improve the robustness of neural networks. Numerous works have been proposed to generate adversarial examples \cite{goodfellow2014explaining,madry2017towards,andriushchenko2020square}. Gradient-based attack methods are commonly used \cite{aldahdooh2022adversarial}. Goodfellow et al. \cite{goodfellow2014explaining} propose to quickly compute adversarial training examples by using the gradient sign. Madry et al. \cite{madry2017towards} propose projected gradient descent (PGD) to compute the adversarial training samples. PGD executes an iterative computation that performs multiple gradient descent updates with small steps within the perturbation bound $\epsilon$ to update the adversarial training samples.





\section{Methodology}
This section first discusses several typical learning methods related to logit perturbation.
\subsection{Logit Perturbation in Existing Methods}
The notations and symbols are defined as follows. Let $\displaystyle S = \{\bm x_i, \bm y_i\}_{i=1}^N$ be a corpus of $N$ training samples, where $\bm x_i$ is the input feature and $\bm y_i$ is the label. In single-label classification, $\bm y_i$ is a one-hot vector. In multi-label classification, ${\bm y_i}=\left[ {{y_{i,1}},{y_{i,2}},\cdots ,{y_{i,C}}} \right] \in {\left\{ {0,1} \right\}^C}$. Let $C$ be the number of categories and $\pi_c{\rm{ }} = {\rm{ }}N_c/N$ be the proportion of the samples, where $N_c$ is the number of the samples that contain the $c$th category in $S$. Without loss of generality, we assume that $\pi_1 > {\rm{ }} \cdots  > {\rm{ }}\pi_c > {\rm{ }} \cdots  > \pi_C$.
Following Menon et al.~\cite{menon2020long} and Wu et al. \cite{wu2020dist}, we determine the head and tail categories by $N_c$. The larger $N_c$ means that $c$ is the head category index, and the smaller $N_c$ means that $c$ is the tail category index. Following Guo et al. \cite{guo2021long}, if $y_{i,c}=1$, $\bm x_i$ is the positive sample of category $c$; otherwise, $\bm x_i$ is the negative sample of category $c$. Let $\bm u_i$ be the logit vector of $\bm x_i$ which can be obtained by $\bm u_i = f(\bm x_i,\bm W)$, where $f(\cdot,\cdot)$ is the deep neural network with parameter $\bm W$. Let ${\bm \delta _i}$ be the perturbation term of $\bm x_i$. Let $\mathcal{L}$ be the entire training loss and $l_i$ be the loss of $\bm x_i$. The standard cross-entropy (CE) loss is used throughout the study.

\textbf{Logit adjustment (LA)}~\cite{menon2020long}. This method is designed for single-label long-tail classification and achieves competitive performance in benchmark data sets~\cite{wu2021adversarial}. The employed loss in LA is defined as follows:
\begin{equation}
\begin{aligned}
\mathcal{L} & = -\sum\nolimits_i {\log \frac{{\exp ({u_{i,{k}}} + \lambda \log {\pi _{{k}}})}}{{\sum\nolimits_c {\exp ({u_{i,c}} + \lambda \log {\pi _c})} }}},
\end{aligned}\label{LA_pert}
\end{equation}
where $u_{i,k}$ is $k$th element of $\bm u_i$; $y_{i,k}$ is $k$th element of $\bm y_i$; $c$ and $k$ are the category index; and $k$ satisfies $y_{i,k}=1$.
In Eq. (\ref{LA_pert}), the perturbation term $\bm \delta_i$ is as follows:
\begin{equation}
{\bm \delta _i}{\rm{ = }}{\bm {\tilde \delta }}{\rm{ = }}\lambda {{\rm{[log}}{\pi _1}{\rm{,}} \cdots {\rm{,log}}{\pi _c}{\rm{,}} \cdots ,\log {\pi _C}{\rm{]}}^T},
\end{equation}
where $\bm{\tilde \delta}$ is corpus-level vector\footnote{Corpus level is viewed as a special kind of class level in this study.}; $\bm \delta _i$ is sample-level vector; thus ${\bm \delta _i}$ for all the samples in the corpus $\bm S$ are identical. Eq. (\ref{LA_pert}) can be re-written as follows:
\begin{equation}
\mathcal{L}{\rm{ =  - }}\sum\nolimits_i {\log \frac{{\exp ({ u_{i,{k}}})}}{{\sum\nolimits_c {\exp ({ u_{i,c}} + \lambda (\log {\pi _c}{\rm{ - log}}{\pi _{{k}}}))} }}}.
\end{equation}

Previously, we assumed that $\pi_1 > {\rm{ }} \cdots  > {\rm{ }}\pi_c > {\rm{ }} \cdots  > \pi_C $; hence, the losses of the samples in the first category (head) are decreased, while those of the samples in the last category (tail) are increased. The variations of the losses of the rest categories depend on the concrete loss of each sample.

\textbf{Implicitly semantic data augmentation (ISDA)}~\cite{wang2019implicit}. ISDA is an explicit data augmentation method for single-label classification. Given a sample $\bm x_i$, ISDA assumes that each (virtual) new sample can be sampled from a distribution $\mathcal{N}\left( {\bm x_i,{\rm{ }}\bm \Sigma_{k}} \right)$, where $\bm \Sigma_{k}$ is the co-variance matrix for the $k$th category. With the $M$ (virtual) new samples for each sample, the loss becomes
\begin{equation}
\mathcal{L}{\rm{ =  - }}\frac{1}{{N \cdot M}}\sum\nolimits_{i = 1}^N {\sum\nolimits_{m = 1}^M {l(f({\bm x_{i,m},\bm W}),{\bm y_i})} } ,\label{ISDA_pert_ori}
\end{equation}
where $\bm x_{i,m}$ is the $m$th (virtual) new sample for $\bm x_i$. When $M{\rm{ }} \rightarrow {\rm{ }}+\infty $, the upper bound of the loss in Eq. (\ref{ISDA_pert_ori}) becomes
\begin{equation}
\mathcal{L}{\rm{ =  - }}\frac{1}{N}\sum\limits_{i = 1}^N {\log \frac{{\exp ({ u_{i,{k}}})}}{{\sum\limits_{c = 1}^C {\exp ({ u_{i,{c}}} + \frac{\lambda }{2}{{({{\bm{w}}_c} - {{\bm{w}}_{{k}}})}^T}{\bm \Sigma _{{k}}}(\bm{w}_c - {{\bm{w}}_{{k}}}))} }}},\label{ISDA_pert}
\end{equation}
where $c$ and $k$ are the category index and $k$ satisfies $\bm y_{i,k}=1$; $\bm w_c$ is the network parameter for the logit vectors and ${u_{i,c}  = }{{\bm{w}}_{\rm{c}}}^T{\bm{\tilde x_i}}{{ + b_c}}$; $\bm{\tilde x_i}$ is the output of the last feature encoding layer. In contrast with previous data augmentation methods, ISDA does not generate new samples or features.
In Eq. (\ref{ISDA_pert}), there is an implicit perturbation term $\bm {\delta _i}$ defined as follows:
\begin{equation}
\bm {\delta _i}{\rm{ = }}{\bm {\tilde \delta} _{{k}}}{\rm{ = }}\frac{\lambda }{2}\left[ {\begin{array}{*{20}{c}}
{{{({{\bm{w}}_1} - {{\bm{w}}_{{k}}})}^T}{\bm \Sigma _{{k}}}({{\bm{w}}_1} - {{\bm{w}}_{{k}}})}\\
 \vdots \\
{{{({{\bm{w}}_C} - {{\bm{w}}_{{k}}})}^T}{\bm \Sigma _{{k}}}({{\bm{w}}_C} - {{\bm{w}}_{{k}}})}
\end{array}} \right],\label{ISDA_pert2}
\end{equation}
where $\bm {\tilde \delta}_{k}$ is class-level vector; thus $\bm \delta_i$ is the same for each class of samples. Each element of ${\bm \delta _i}$ is non-negative. Therefore, the new loss of each category from Eq. (\ref{ISDA_pert}) is larger than the loss from the standard CE loss.

 \textbf{LDAM}~\cite{cao2019learning}. This method is designed for single-label long-tail classification. It's new loss is defined as
\begin{equation}
\begin{aligned}
&\mathcal{L}=  - \sum\limits_{i = 1}^N {\log \frac{{\exp ({ u_{i,{k}}}{\rm{ - }}C {{({\pi _{{k}}})}^{{\rm{ - }}1/4}})}}{{\exp ({ u_{i,{k}}}{\rm{ - }}C {{({\pi _{{k}}})}^{{\rm{ - }}\frac{1}{4}}}){\rm{ + }}\sum\nolimits_{c \ne {k}} {\exp ({ u_{i,c}})} }}},
\end{aligned}\label{LDAM_loss}
\end{equation}
where  $k$ satisfies $ y_{i,k}=1$.
The perturbation term ${\bm \delta _i}$ is as follows:
\begin{equation}
{\bm \delta _i}{\rm{ = }}{\bm {\tilde \delta} _{{k}}}{\rm{ = }}\lambda {{\rm{[}}0{\rm{ , }} \cdots {\rm{, - }}C{({\pi _{{k}}})^{{\rm{ - }}\frac{1}{4}}}{\rm{,}} \cdots , 0{\rm{]}}^T},\label{LDAM_pert}
\end{equation}
which is also a category-level vector. The losses for all categories are increased in LDAM. 

\begin{figure*}[h] 
\centering
\includegraphics[width=1.0\textwidth,height=0.16\textwidth]{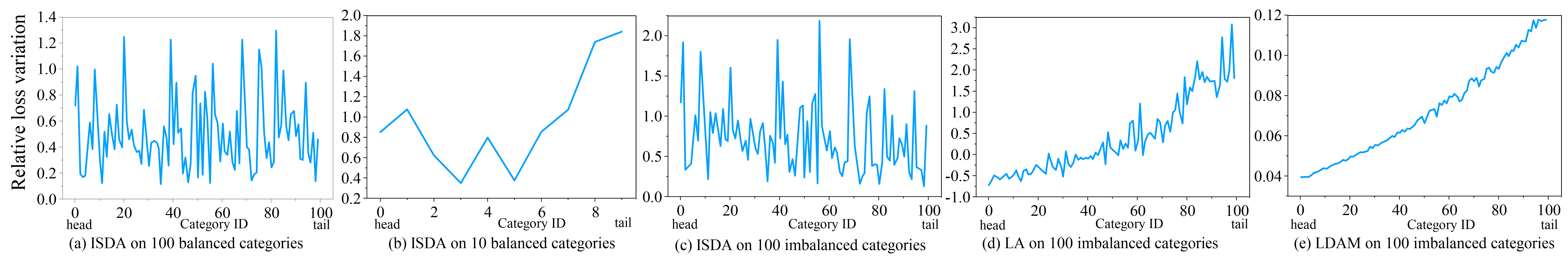} \vspace{-0.3in}
\caption{The relative loss variations ($\frac{l'-l}{l}$) of the three methods on different categories on different data sets. (a) and (b) show the relative loss variation of ISDA on CIAFR100 and CIFAR10 respectively. (c), (d) and (e) show the relative loss variation of ISDA, LA and LDAM on CIAFR100-LT with imbalance ratio 100:1, respectively.}
\label{relative_loss_variations_single}
\end{figure*}

\begin{figure}[b]
\centering
\includegraphics[width=0.9\columnwidth, height=1.1in]{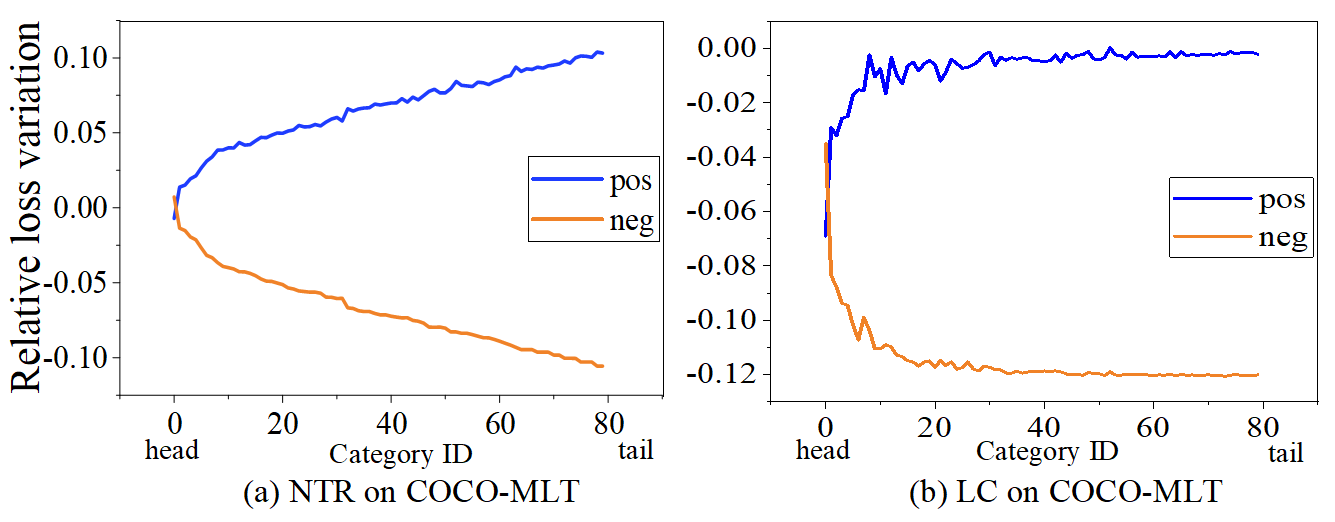}  \vspace{-0.15in}
\caption{The relative loss variations ($\frac{l'-l}{l}$) of the two methods on different categories on COCO-MLT. ``pos" means the relative loss variations of positive samples. ``neg" means the relative loss variations of negative samples. }
\label{relative_loss_variations_multi}
\end{figure}



\textbf{Negative-tolerant Regularization (NTR)}~\cite{wu2020dist}. In this method, a multi-label classification task is first decomposed into $C$ independent binary classification tasks. NTR defines the following negative-tolerant binary loss:
\begin{equation}
\begin{aligned}
\mathcal{L}&= \frac{1}{N}\sum_{i=1}^N \frac{1}{C}\sum_{c=1}^C y_{i,c}\text{log}(1+\exp(- u_{i,c}+v_c)) \\ 
&+\frac{1}{\lambda}(1- y_{i,c})\text{log}(1+\exp(\lambda( u_{i,c}-v_c)))
\end{aligned},\label{NTR_loss}
\end{equation}
where $v_c =\psi  \text{log}(\frac{N}{N_c}-1) $; $\lambda$ and $\psi $ are hyper-parameters.
In Eq.~(\ref{NTR_loss}) , an implicit logit perturbation term ($\bm \delta_i$) can also be observed as follows:
\begin{equation}
\bm {\delta _i}{\rm{ = }}{\bm {\tilde \delta }}{\rm{ = }}-\psi [\text{log}(\frac{N}{N_1}-1) ,\cdots,(\frac{N}{N_C}-1) ]^T.\label{NTR_pert}
\end{equation}
The perturbation is a corpus-level term vector. $\psi$ is non-negative in the experiments conducted by Wu et al.~\cite{wu2020dist}. Therefore, if $N < 2N_c$, then samples with label $c$ are dominant and $v_c$ in Eq.~(\ref{NTR_loss}) is smaller than zero. When $ y_{i,c}$=1, the loss will be reduced, and if $ y_{i,c}$=0, the loss will be increased. When $N > 2N_c$, it is opposite.

\textbf{Logit Compensation (LC)}~\cite{guo2021long}. LC assumes that logits conform to a normal distribution. The loss of logit compensation is defined as follows:
\begin{equation}
\begin{aligned}
\mathcal{L}&= \frac{1}{N}\sum_{i=1}^N \frac{1}{C}\sum_{c=1}^C y_{i,c}\text{log}(1+\exp(-(\sigma_c^p\cdot  u_{i,c} + \mu_c^p))) \\ 
&+(1-y_{i,c})\text{log}(1+\exp({\sigma_c^n\cdot  u_{i,c}} + \mu_c^n))
\end{aligned},
\end{equation}
where $\mu_c^p$, $\sigma_c^p$, $\mu_c^n$, and $\sigma_c^n$ ($c\in \{1, \cdots, C\}$) are the mean and variance of the positive and negative samples that can be learned. For the positive samples, the perturbation term $\delta_i$ is as follows:
\begin{equation}
\bm {\delta _i}{\rm{ = }}\bm {\tilde \delta}{\rm{ = }}[ {\mu _1^p,\mu _2^p, \cdots ,\mu _C^p} ].
\end{equation}
For the negative samples, the perturbation term $\delta_i$ is as follows:
\begin{equation}
\bm {\delta _i}{\rm{ = }}\bm {\tilde \delta}{\rm{ = }}[ {\mu _1^n,\mu _2^n, \cdots ,\mu _C^n} ].
\end{equation}
Both the two perturbation items are corpus-level vectors.
In addition, the logit is weighted in accordance with the variance simultaneously. According to the analysis in \cite{guo2021long}, LC mainly (relatively) increases the loss for positive samples and emphasizes the tail categories.

\subsection{Theoretical Analysis for Logit Perturbation}
The losses of the five example methods analyzed in the previous subsection can be written as follows:
\begin{equation}
\mathcal{L}{\rm{ = }}\sum\nolimits_i {l(\bm {u_i} + {\bm {\tilde \delta }_{i}},\bm {y_i})}. \label{common_loss}
\end{equation}
Logit perturbations result in the loss variations. Fig. \ref{relative_loss_variations_single} shows the statistics for the relative loss variations incurred by ISDA, LA, and LDAM for each category on a balanced data set (CIFAR100 \cite{krizhevsky2009learning}) and two long-tail sets (CIFAR10-LT \cite{cui2019class} and CIFAR100-LT \cite{cui2019class}) which are introduced in the experimental section. The loss variations of all categories are positive using ISDA. ISDA achieves the worst results on CIFAR100-LT\cite{cui2019class} (shown in the experimental parts), indicating that the non-tail-priority augmentation in long-tail problems is ineffective (ISDA achieves relatively better results on CIFAR10-LT\cite{cui2019class}.). Only the curves on CIFAR100-LT are shown for LA and LDAM because similar trends can be observed on CIFAR10-LT. The loss variations of head categories are negative, and those of tail are positive using LA. All the variations are positive yet there is an obvious increasing trend using LDAM. Fig. \ref{relative_loss_variations_multi} shows the statistics for the relative loss variations incurred by NTR and LC in multi-label classification. The data set COCO-MLT\cite{wu2020dist} is used. The relative loss variations of positive samples and negative samples are counted separately. In NTR, for positive samples, the loss variations of head categories are less than 0, and those of tail categories are greater than 0. However, for negative samples,  the situation is opposite. LC and NTR have the similar trend of the relative loss variation, but the relative loss variation of LC is less than 0.


We propose two conjectures based on the above observations and from a unified logit-perturbation data augmentation viewpoint: 
\begin{itemize}
\item If one aims to positively augment the samples in a category, the training loss of this category should be increased after logit perturbation. The larger the loss increment is, the greater the augmentation will be. Consequently, the performance of this category will (relatively) increase.
\item If one aims to negatively augment the samples in a category, then the training loss of this category should be reduced after logit perturbation. The larger the loss decrement is, the greater the negative augmentation will be. The performance of this category will (relatively) decrease.
\end{itemize}

The above two conjectures are empirically supported by the aforementioned five methods. For single-label classification, to handle a long-tail problem, LA should positively augment tail samples and negatively augment head samples. Hence, the losses of tail samples are increased, and those of heads are decreased. ISDA aims to positively augment samples in all categories; thus, the losses for all categories are increased. LDAM aims to positively augment tail samples more than head samples. Hence, the increments of tail categories are larger than those of the head. For multi-label classification task, positive samples and negative samples need to be considered separately. For positive samples, NTR positively augments the tail categories and negatively augments the head categories. For negative samples, the condition is opposite. Therefore, the losses of tails are increased, whereas those of heads are decreased. LC aims to negatively augment all categories. For positive samples, the reductions of head categories are larger than those of the tail. For negative samples, the situation is opposite.

To theoretically support the two conjectures, a simple binary classification task is employed to quantitatively investigate the relationship among loss variations, performance improvement, and logit perturbation. The binary classification setting established by Xu et al.~\cite{xu2021robust} is followed. The data from each of the two classes $\displaystyle{\mathcal{Y}}=$ $\{-1, +1\}$  follow two Gaussian distributions, which are centered on $\boldsymbol{\theta} = [\eta, \cdots,\eta]$ ($d$-dimensional vector and $\eta > 0$) and $-\boldsymbol{\theta}$, respectively.
The data follow
\begin{equation} 
y \stackrel{u . a . r}{\sim}\{-1,+1\},
\end{equation}
\begin{equation}
\bm x \sim\left\{\begin{array}{ll}\mathcal{N}\left(\boldsymbol{\theta}, \sigma_{+}^{2} \boldsymbol{I}\right) & \text { if } y=+1 \\ \mathcal{N}\left(-\boldsymbol{\theta}, \sigma_{-}^{2} \boldsymbol{I}\right) & \text { if } y=-1\end{array}\right.\label{data_distribution}.
\end{equation}
For a classifier $f$, the overall standard error is defined as $\mathcal{R}(f)=\operatorname{Pr}.(f(\bm x)\neq y)$. We use $\mathcal{R}(f;y)$ to denote the standard error conditional on a specific class $y$. The class ``+1" is harder because an optimal linear classifier will give a larger error for the class ``+1" than that for the class ``-1" when $\sigma_{+}^{2} > \sigma_{-}^{2}$ \cite{xu2021robust}.
Two types of class-level logit perturbation are considered in our theoretical analysis. Let $\epsilon_c$ be the perturbation bound. The first type of perturbation is defined as follows:
\begin{equation}
{{\tilde \delta }_{{c}}}^*{\rm{ = }}arg \max_{||{{\tilde \delta }_{{c}}}|| < \epsilon_c} \text{E}_{(\bm x,y):y=c} [{l({u} + {{\tilde \delta }_{{c}}},{c})}] \label{first_perturbation}.
\end{equation}
The second type is defined as follows:
\begin{equation}
{{\tilde \delta }_{{c}}}^*{\rm{ = }}arg \min_{||{{\tilde \delta }_{{c}}}|| < \epsilon_c} \text{E}_{(\bm x,y):y=c}[ {l({u} + {{\tilde \delta }_{{c}}},{c})}], \label{second_perturbation}
\end{equation}
where $ u=\boldsymbol{w}^T\bm x+b$.
The first type implements positive augmentation, while the second type implements negative augmentation.

Assuming that the perturbation bounds between the two classes satisfy that $\epsilon _+ =\rho_+\cdot\epsilon$ and $\epsilon _- = \rho_-\cdot\epsilon$. Now, the variances of the data distributions in Eq. (\ref{data_distribution}) for the two classes are assumed to be equal, i.e., $\sigma_{+} = \sigma_{-}$. Nevertheless, the prior probabilities of the two classes $P(y=+1)$ ($P_+$) and $P(y=-1)$ ($P_-$) are assumed to be different. Without loss of generality, we assume $P_{+}: P_{-}=1: \Gamma $ and $ \Gamma >1$. That is, class imbalance exists, and the class $+1$ and the class $-1$ are the minority and the majority classes, respectively.
We have the following theorem:
\begin{thm}
For the abovementioned binary classification task,  the logit perturbation bounds of classes  ``$+1$" and ``$-1$" are assumed to be $\rho_+\cdot\epsilon$ ($0\leq\rho_+\cdot{\epsilon}<{\eta}$) and $\epsilon$ ($\rho_- =1$), respectively. Only the first perturbation type is utilized. The optimal linear classifier $f_{\text{opt}}$ that minimizes the average classification error is 
\begin{equation}
   f_{\text{opt}}=\arg\underset{f}{ \min } \operatorname{Pr}.(\mathbb{S}(u+{{\tilde \delta }_{{c}}}^*) \neq y), \label{optf_thm1}
\end{equation}
where $u=f(\bm x)=\boldsymbol{w}^T\bm x+b$;  $\mathbb{S}(\cdot)$ is the signum function (if $a \geq 0$, then $\mathbb{S}(a) = 1$; else $\mathbb{S}(a) = -1$).
It has the intra-class standard error for the two classes:
\begin{equation}
\small
\begin{aligned} 
&\mathcal{R}\left(f_{\text{rob}},-1\right) = \operatorname{Pr}.\left\{\mathcal{N}(0,1)<\frac{A}{2}+\frac{\text{log}\Gamma }{A}-\frac{\epsilon}{\sqrt{d}\sigma}\right\}, \\ & \mathcal{R}\left(f_{\text{rob}},+1\right)= \operatorname{Pr}.\left\{\mathcal{N}(0,1) <\frac{A}{2}-\frac{\text{log}\Gamma }{A}-\frac{\epsilon\rho_+}{\sqrt{d}\sigma}\right\},
    \end{aligned}
\end{equation}
where $A=\frac{\epsilon-2d\eta+\epsilon\rho_+}{\sqrt{d}\sigma}$.\label{imbal_thm1}
\end{thm}
The proof is attached in the appendix. Theorem 1 indicates that the logit perturbation parameterized by $\epsilon$ and $\rho_+$ influences performance of both classes. We then show how the classification errors of the two classes change as $\rho_+$ increases.

\begin{figure}[tb]
\centering
\includegraphics[width=0.95\columnwidth, height=1.1in]{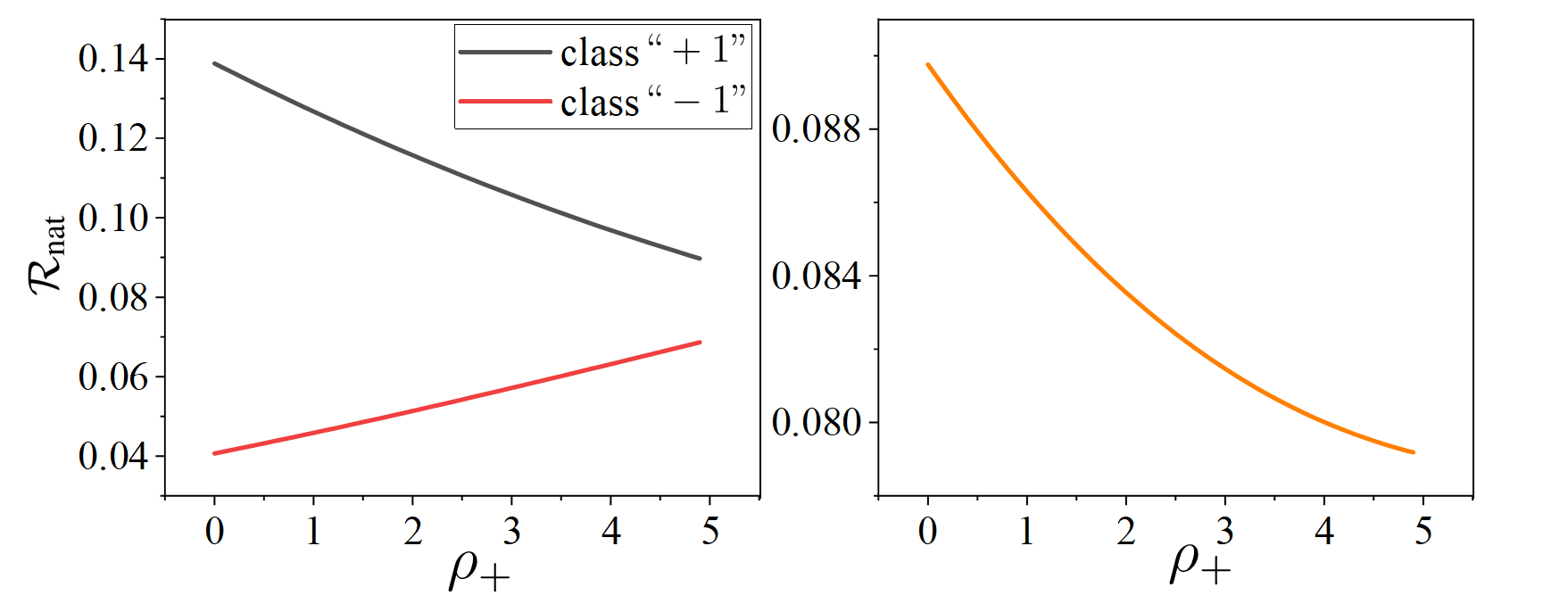}  \vspace{-0.15in}
\caption{Left: Natural errors $\mathcal{R}\left(f_{\text{opt}},-1\right)$ and $\mathcal{R}\left(f_{\text{opt}},+1\right)$ for the two classes with varied $\rho_+$. Right: Total natural error $\mathcal{R}(f_{\text{opt}})$ with
varied $\rho_+$.}
\label{imbal_lpl1_fig}
\end{figure}

\begin{corollary}
For the binary classification task investigated in Theorem 1, when $\Gamma<e^{\frac{((2d-1)\eta-\epsilon)^2}{2d\sigma^2}}$, as $\rho_+$ increases, the logit perturbations on Theorem 1 will decrease the error for class ``$+1$" and increase the error for class ``$-1$".\label{imbal_coro1}
\end{corollary}
The proof is attached in the appendix. Corollary 1 indicates that a larger scope of the first type of logit perturbation on a class will increase the performance of the class. Note that a larger scope of the first type of logit perturbation will result in a large loss increment, and the first conjecture is supported by Corollary 1. To better illuminate Corollary \ref{imbal_coro1}, we plot $\mathcal{R}\left(f_{\text{opt}},-1\right)$, $\mathcal{R}\left(f_{\text{opt}},+1\right)$, and $\mathcal{R}(f_{\text{opt}})$ for a specific learning task. Fig. \ref{imbal_lpl1_fig} shows the results when the values of $\Gamma$, 
$d$, $\eta$, $\epsilon$, and $\sigma$ are 2, 2, 1, 0.2, and 1, respectively.

Theorem 1 only considers the first type of logit perturbation. When the second type of logit perturbation is also involved, the following theorem can be obtained.

\begin{thm}
For the abovementioned binary classification task, that the perturbation bounds of both classes ``$+1$" and ``$-1$" are assumed to be $\epsilon$ ($\rho_+=1$)  and $\rho_-\cdot\epsilon$ ($0\leq\rho_-\cdot{\epsilon}<{\eta}$), respectively. The first perturbation type is utilized for class ``$+1$", and the second perturbation type is utilized for ``$-1$". The optimal linear classifier $f_{\text{opt}}$ that minimizes the average classification error is 
\begin{equation}
   f_{\text{opt}}=\arg\underset{f}{ \min } \operatorname{Pr}(\mathbb{S}(u+{{\tilde \delta }_{{c}}}^*) \neq y). 
\end{equation}
It has the intra-class standard error for the two classes:
\begin{equation}
\begin{aligned} & \mathcal{R}\left(f_{\text{opt}},-1\right) = \operatorname{Pr}.\left\{\mathcal{N}(0,1)<\frac{A}{2}+\frac{\text{log}\Gamma}{A}+\frac{\epsilon\rho_-}{\sqrt{d}\sigma}\right\}, \\ & \mathcal{R}\left(f_{\text{opt}},+1\right) = \operatorname{Pr}.\left\{\mathcal{N}(0,1) <\frac{A}{2}-\frac{\text{log}\Gamma}{A}-\frac{\epsilon}{\sqrt{d}\sigma}\right\}, \end{aligned}
\end{equation}
where $A=\frac{\epsilon-2d\eta-\epsilon\rho_-}{\sqrt{d}\sigma}$.
\label{imbal_thm2}
\end{thm}
The proof of Theorem \ref{imbal_thm2}  is similar to that of Theorem \ref{imbal_thm1}.  Likewise, we have the following corollary.
\begin{corollary}
For the learning task investigated in Theorem 2, when $\Gamma>1$, as $\rho_-$ increases, the logit perturbations on Theorem \ref{imbal_thm2} will increase the accuracy for class ``$+1$" and decrease the accuracy for class ``$-1$".\label{imbal_coro2}
\end{corollary}

\begin{figure}[tb]
\centering
\includegraphics[width=0.95\columnwidth, height=1.1in]{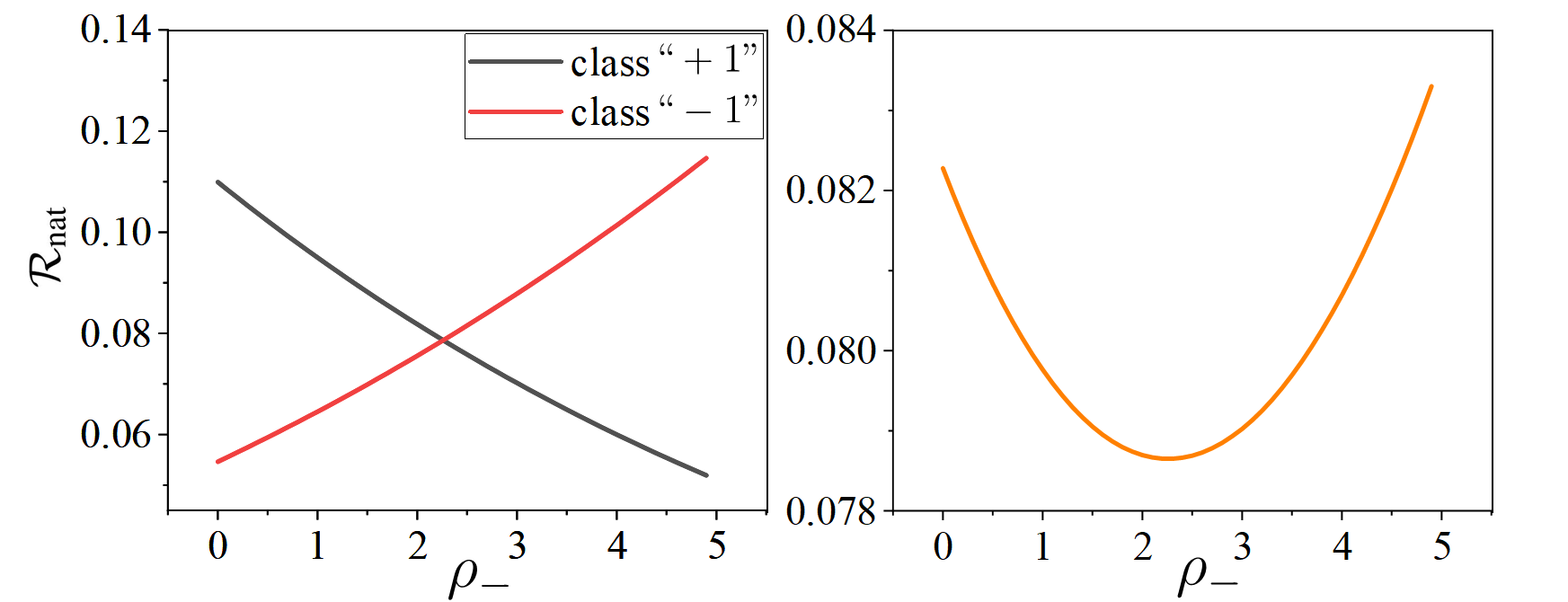}  \vspace{-0.15in}
\caption{Left: Natural errors $\mathcal{R}\left(f_{\text{opt}},-1\right)$ and $\mathcal{R}\left(f_{\text{opt}},+1\right)$ for the two classes with varied $\rho_-$. Right: Total natural error $\mathcal{R}(f_{\text{opt}})$ with
varied $\rho_{-}$.}
\label{imbal_lpl2_fig}
\end{figure}

According to Corollary 2, a larger scope of the second type of logit perturbation
on a class will decrease the performance of the class. Note
that a larger scope of the second type of logit perturbation
will result in a large loss decrement, and the second conjecture is supported.   Likewise, we plot $\mathcal{R}\left(f_{\text{opt}},-1\right)$, $\mathcal{R}\left(f_{\text{opt}},+1\right)$, and $\mathcal{R}(f_{\text{opt}})$. Fig. \ref{imbal_lpl2_fig} shows the results for the specific learning task discussed in Fig. \ref{imbal_lpl1_fig}. In this figue, the values of $\Gamma$, 
$d$, $\eta$, $\epsilon$, and $\sigma$ are 2, 2, 1, 0.2, and 1, respectively.

Theorems 1-2 and Corollaries 1-2 concern the class imbalance issue, i.e., $P_{+} \neq P_{-}$. In addition, another learning scenario is also explored. The variances of the data distributions in Eq. (\ref{data_distribution}) for the two classes are assumed to be unequal, i.e., $\sigma_{+} \neq \sigma_{-}$. That is, variance imbalance exists. Without loss of generality, we assume $\sigma_{+} : \sigma_{-} = 1:K$, where $K> 1$. And $P_{+}: P_{-}=1: \Gamma $ also holds, where $\Gamma > 1$.
We have the following theorem.
\begin{thm}
For the abovementioned binary classification task, the bounds of classes  ``$+1$" and ``$-1$" are assumed to be $\rho_+\cdot{\epsilon}$  and $\rho_-\cdot{\epsilon}$ ($0\leq \rho_+, \rho_-<\frac{\eta}{\epsilon}$), respectively. Only the first perturbation type is utilized. The optimal linear classifier $f_{\text{opt}}$ that minimizes the average classification error is 
\begin{equation}
   f_{\text{opt}}=\arg\underset{f}{ \min } \operatorname{Pr}.(\mathbb{S}(u+{{\tilde \delta }_{{c}}}^*) \neq y), \label{optf}
\end{equation}
where $u=f(x)=\boldsymbol{w}^T\bm x+b$.
It has the intra-class standard error for the two classes:
\begin{equation}
\small
\begin{aligned} & \mathcal{R}\left(f_{\text{opt}},+1\right)
\\&=\operatorname{Pr}.\left\{\mathcal{N}(0,1)<-K\sqrt{B^2+q(K,\Gamma)}-B-\frac{\epsilon\cdot\rho_+}{\sqrt{d}\sigma})\right\}, \\ & \mathcal{R}\left(f_{\text{opt}},-1\right) \\&= \operatorname{Pr}.\left\{\mathcal{N}(0,1)<KB+\sqrt{B^2+q(K,\Gamma)}-\frac{\epsilon\cdot\rho_-}{K\sqrt{d}\sigma}\right\}, \end{aligned}
\end{equation}
where $B= \frac{\epsilon\cdot\rho_++\epsilon\cdot\rho_--2d\eta}{\sqrt{d}\sigma(K^2-1)}$ and $q(K,\Gamma)=\frac{2\text{log}(\frac{K}{\Gamma})}{K^2-1}$.\label{Bal_thm1}
\end{thm}

Thus, training with different logit perturbation bounds for the two classes can still influence the performance according to Theorem \ref{Bal_thm1}. We then show how the classification errors of the two classes change as $\rho_-$ or $\rho_+$ increases.
\begin{corollary}
For the data distribution and logit perturbation investigated in Theorem 3,
\begin{itemize}
    \item if  $Ke^{\frac{(2d\eta-\epsilon)^2}{2dK^2\sigma^2}}<\Gamma< Ke^{\frac{2d\eta^2}{(K^2-1)\sigma^2}}$, then $\mathcal{R}\left(f_{\text{opt}},+1\right)>\mathcal{R}\left(f_{\text{opt}},-1\right)$. That is, class imbalance is the primary challenge and class ``$+1$'' is harder than class ``$-1$''. Then if $\rho_- = 1$ and the first logit perturbation type is used, the error of class ``$+1$'' decreases and the error of class ``$-1$'' increases, as $\rho_+$ increases;
    \item if $K>\Gamma$, then $\mathcal{R}\left(f_{\text{opt}},+1\right)<\mathcal{R}\left(f_{\text{opt}},-1\right)$. That is,  variance imbalance is the primary challenge and class ``$-1$'' is harder than class ``$+1$''. If $\rho_+ = 1$ and the first logit perturbation type is used, the error of class ``$+1$'' increases and the error of class ``$-1$'' decreases, as $\rho_-$ increases.
    
\end{itemize}  
 \label{bal_coro1}
\end{corollary}



\begin{figure}[tb]
\centering
\includegraphics[width=0.95\columnwidth, height=1.1in]{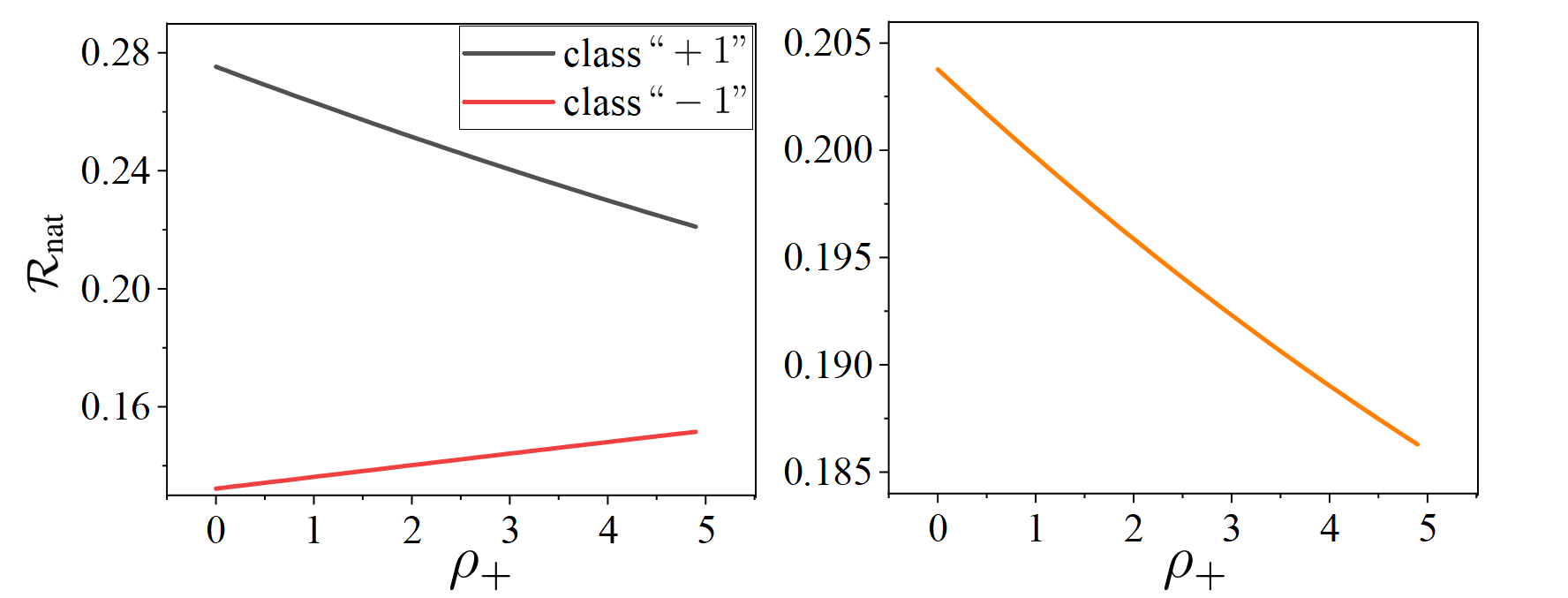}  \vspace{-0.15in}
\caption{Left: Natural errors $\mathcal{R}\left(f_{\text{opt}},-1\right)$ and $\mathcal{R}\left(f_{\text{opt}},+1\right)$ for the two classes with varied $\rho_+$. Right: Total natural error $\mathcal{R}(f_{\text{opt}})$ with
varied $\rho_+$.}
\label{bal_lpl_fig1}
\end{figure}

\begin{figure}[tb]
\centering
\includegraphics[width=0.95\columnwidth, height=1.1in]{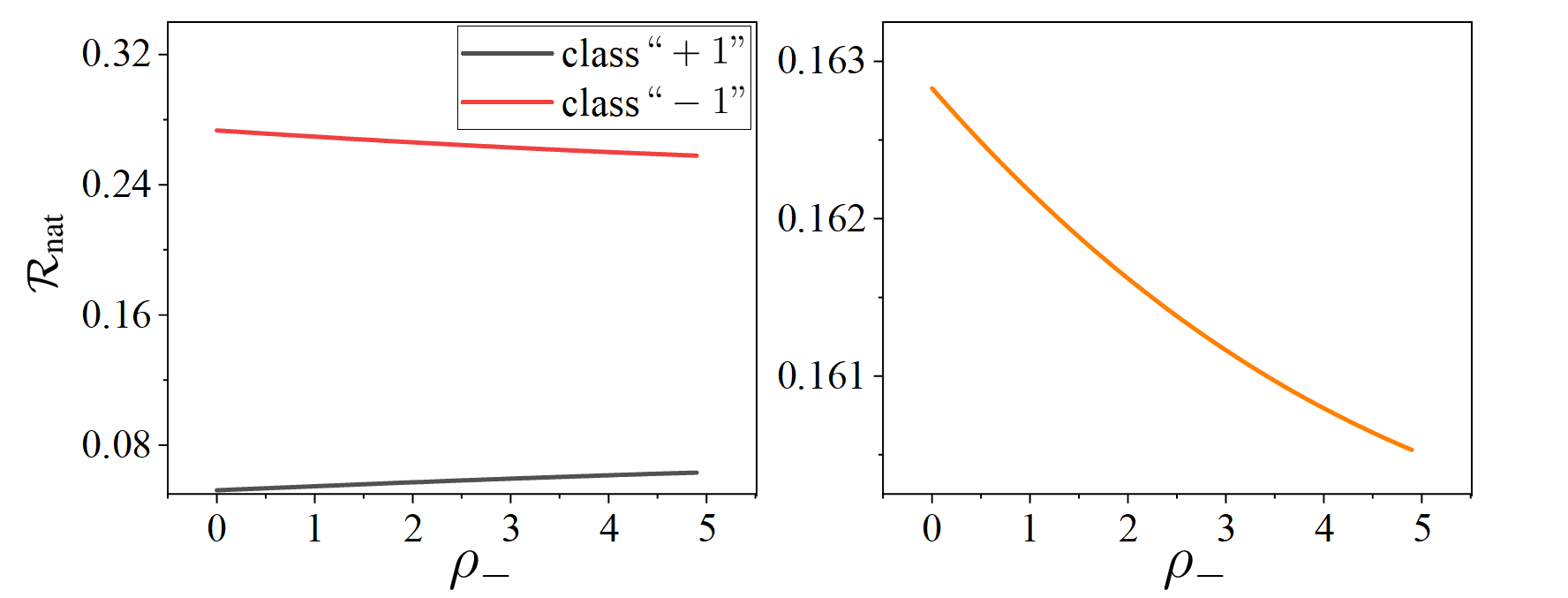}  \vspace{-0.15in}
\caption{Left: Natural errors $\mathcal{R}\left(f_{\text{opt}},-1\right)$ and $\mathcal{R}\left(f_{\text{opt}},+1\right)$ for the two classes with varied $\rho_-$. Right: Total natural error $\mathcal{R}(f_{\text{opt}})$ with
varied $\rho_-$.}
\label{bal_lpl_fig2}
\end{figure}

The first conjecture can also be justified by Corollary \ref{bal_coro1}. Likewise, we plot $\mathcal{R}\left(f_{\text{opt}},-1\right)$, $\mathcal{R}\left(f_{\text{opt}},+1\right)$, and $\mathcal{R}(f_{\text{opt}})$. Figs. \ref{bal_lpl_fig1} and \ref{bal_lpl_fig2} show the results. As shown in Fig. \ref{bal_lpl_fig1}, increasing $\rho_+$ can decrease the error of class ``$+1$'' and increase the error of class ``$-1$''. The values of $K$, $\Gamma$,
$d$, $\eta$, $\epsilon$, and $\sigma$ are 3, 3.5, 2, 1, 0.1 and 1, respectively. In  Fig. \ref{bal_lpl_fig2}, increasing $\rho_-$ can decreases the error of class ``$-1$'' and increase the error of class ``$+1$''. The values of $K$, $\Gamma$,
$d$, $\eta$, $\epsilon$, and $\sigma$ are 2.5, 1.1, 2, 1, 0.2 and 1, respectively.

When both types are utilized, we can obtain the following conclusion. When class ``$-1$'' is harder than class ``$+1$'', if the first logit perturbation type is used for class ``$-1$'' and the second logit perturbation is used for class ``$+1$'', then the error of class ``$-1$'' will decrease and the error of class ``$+1$'' will increase. Similarly, when class ``$+1$'' is harder than class ``$-1$'', if the first logit perturbation is used for class ``$+1$'' and the second logit perturbation type is used for class ``$-1$'', then the error of class ``$+1$'' will decrease and the error of class ``$-1$'' will increase. That is, the second conjecture is also justified.

\subsection{Logit Perturbation Method (LPL) for Single-label Learning}
\begin{figure}[t]
\centering
\includegraphics[width=0.96\columnwidth, height=1.1in]{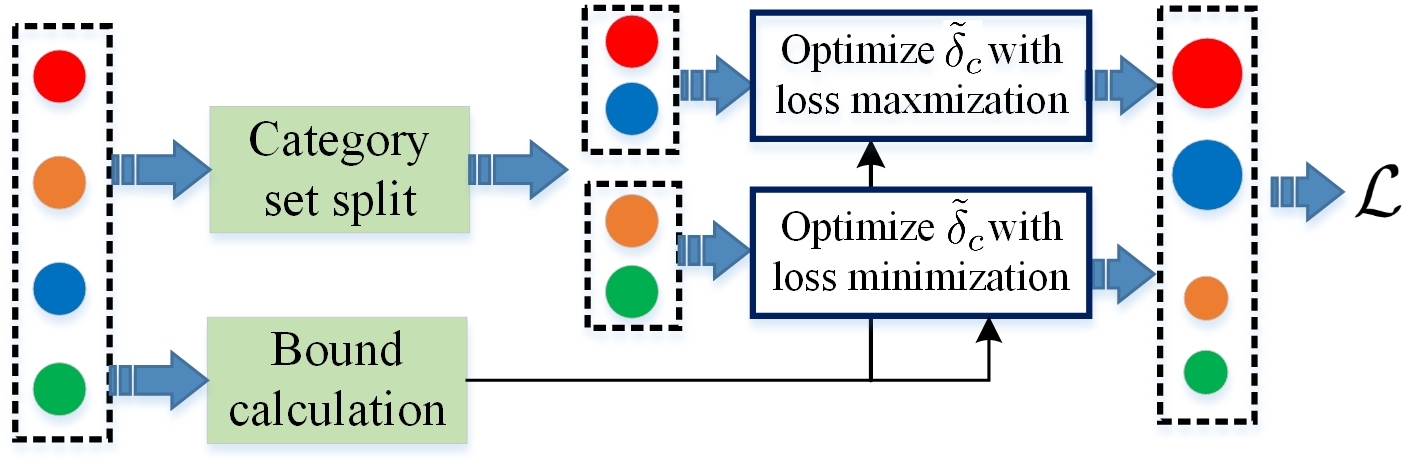} \vspace{-0.1in}
\caption{Overview of the logit perturbation-based new loss. Four solid circles denote four categories. Two categories are positively augmented via loss maximization and the rest two are negatively augmented via minimization.}
\label{LPL_illustrate}
\end{figure}

On the basis of our conjectures and theoretical investigation, we establish the following new training loss with logit perturbation:
\begin{equation}
\begin{aligned}
\mathcal{L}{\rm{ = }}\sum\limits_{{c} \in {\displaystyle {\mathcal{N}}_a}} {\sum\limits_{\bm {x}_i \in \displaystyle{S_c}} {\mathop {\min }\limits_{\left\| {\bm {\tilde \delta _{{c}}}} \right\| \le \epsilon_c } l(\text{softmax} (\bm {u}_i + {\bm {\tilde \delta }_{{c}}}),\bm {y}_i)} } \\
{\rm{        + }}\sum\limits_{{c} \in \displaystyle {\mathcal{P}_a}} {\sum\limits_{\bm {x_i} \in \displaystyle{S_c}} {\mathop {\max }\limits_{\left\| {\bm {\tilde \delta _{{c}}}} \right\| \le \epsilon_c } l(\text{softmax} (\bm {u}_i + {\bm {\tilde \delta }_{{c}}}),\bm {y}_i)} },
\end{aligned}\label{new_loss_abs}
\end{equation}
where ${\epsilon _c}$ is the perturbation bound related to the extent of augmentation; $\displaystyle {\mathcal{N}}_a$ is the index set of categories which should be negatively augmented; $\displaystyle {\mathcal{P}}_a$ is the index set of categories which should be positively augmented; and $\displaystyle {S}_c$ is the set of samples in the $c$th category. The loss maximization for the $\displaystyle {\mathcal{P}}_a$ categories is actually the category-level adversarial learning on the logits; the loss minimization for the $\displaystyle{\mathcal{N}}_a$ categories is the opposite. Fig. \ref{LPL_illustrate} illustrates the calculation of the logit perturbation-based new loss in Eq. (\ref{new_loss_abs}).

The split of the category set (i.e., $\displaystyle{\mathcal{N}}_a$ and $\displaystyle{\mathcal{P}}_a$) and the definition (calculation) of $\epsilon_c$ are crucial for the learning with Eq. (\ref{new_loss_abs}). Category set split determines the categories that should be positively or negatively augmented. Meanwhile, the value of $\epsilon_c$ determines the augmentation extent.



\textbf{Category set split}. The split depends on specific learning tasks. Two common cases are explored in this study. The first case splits categories according to their performances. In this case, Eq. (\ref{new_loss_abs}) becomes the following compact form:
\begin{equation}
\begin{aligned}
\mathcal{L} &=\sum\nolimits_c\{\mathbb{S}(\tau- {{\bar q}_{c}}) \times \\
&\sum\limits_{\bm {x_i} \in \displaystyle{S}_c} {\mathop {\max }\limits_{\left\| {{\bm {\tilde \delta }_{{c}}}} \right\| \le \epsilon_c } [ l(\text{softmax} ({\bm u_i} + {\bm {\tilde \delta }_{{c}}}),\bm {y}_i)\mathbb{S}(\tau  - {{\bar q}_{c}})]\} } ,
\end{aligned}\label{special_loss1}
\end{equation}
where $\tau$ is a threshold, $y_{i,c} = 1$,  and $\bar q_{c}$ is calculated by
\begin{equation}
{\bar q_{c}}{\rm{ = }}\frac{1}{{{\rm{}}{N_{{c}}}{\rm{}}}}\sum\limits_{\bm x_i \in {\displaystyle S_{{c}}}} {{q_{i,{c}}}}  = \frac{1}{{{\rm{}}{N_{{c}}}{\rm{}}}}\sum\limits_{\bm x_i \in {\displaystyle S_{{c}}}} {\frac{{\exp ({u_{i,{c}}})}}{{\sum\nolimits_{c'} {\exp ({ u_{i,c'}} )} }}}. \end{equation}

When $\tau {\rm{ }} = \text{mean}({\bar q_{c}})=\sum\nolimits_{c=1}^C{\bar q_{c}}/C$, Eq. (\ref{special_loss1}) indicates that if the performance of a category is below the mean performance, it will be positively augmented. Meanwhile, when the performance is above the mean, it will be negatively augmented. When $\tau {\rm{ }} > \max \limits_c\{\bar q_{c}\}$, all the categories will be positively augmented as in ISDA.
\begin{figure}[tb]
\centering
\includegraphics[width=1\columnwidth, height=1.1in]{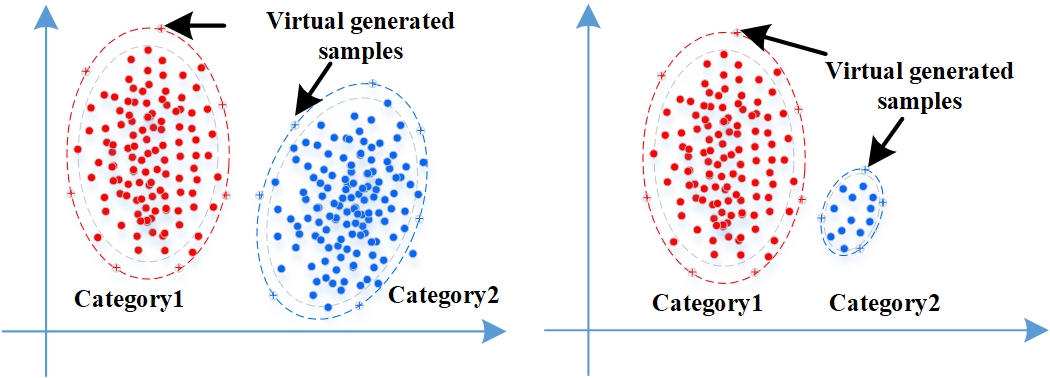} \vspace{-0.15in}
\caption{Illustrative example for ISDA. Both categories are positively augmented (new samples are virtually generated) according to feature distributions.}
\label{fig_ISDA}
\end{figure} 

\begin{figure}[tb]
\centering
\includegraphics[width=1\columnwidth, height=1.1in]{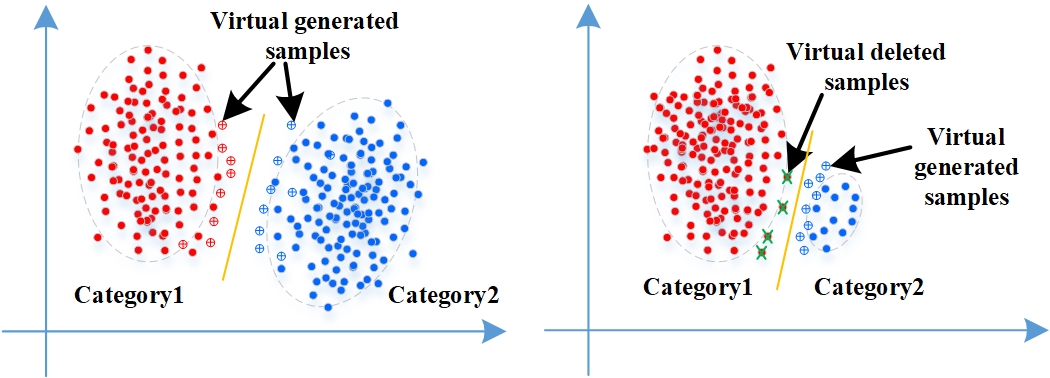}  \vspace{-0.15in}
\caption{Illustrative example for LPL. Samples near the classification boundary are virtually generated or deleted.}
\label{fig_LPL}
\end{figure}

The second case is special for a long-tail problem, and it splits categories according to the proportion order of each category. Eq. (\ref{new_loss_abs}) becomes the following compact form:
\begin{equation}
\begin{aligned}
\mathcal{L}&=\sum\nolimits_c\{\mathbb{S}(c - \tau ) \times \\
&\sum\limits_{{\bm x_i} \in {\displaystyle S_c}} {\mathop {\max }\limits_{\left\| {{\bm{\tilde \delta }_{{c}}}} \right\| \le \epsilon_c } [ l(\text{softmax} ({\bm u_i} + {\bm {\tilde \delta }_{{c}}}),\bm y_i)\mathbb{S}(c - \tau )]\} },
\end{aligned}\label{longtail_new_loss}
\end{equation} 
where $\tau$ is a threshold for the category index and $ y_{i,c} = 1$. In Eq. (\ref{longtail_new_loss}), the tail categories locate in $\displaystyle {\mathcal{P}}_a$ and will be positively augmented.

Eqs. (\ref{special_loss1}) and (\ref{longtail_new_loss}) can be solved with an optimization approach similar to PGD~\cite{madry2017towards}. 
We propose a more specific optimization method called PGD-like optimization based on PGD. According to the derivative of the cross-entropy loss function with respect to logit vector, our PGD-like optimization method can be implemented simply. First, we have
\begin{equation}
    \frac{\partial l(\text{softmax}(\bm {u}_{i}+\bm {\tilde{\delta}}_{c}), \bm y_{i})}{\partial \bm {\tilde{\delta}}_{c}}\bigg|_{\boldsymbol{0}}=\text{softmax}(\bm {u}_{i})-\bm {y}_{i}.
\end{equation}
In the maximization of Eqs. (\ref{special_loss1}) and (\ref{longtail_new_loss}), $\bm {\tilde \delta }_{c}$ is updated by

\begin{equation}
\begin{aligned}
\bm {\tilde{ \delta }}_{c} = \frac{\alpha }{N_{c}} \sum_{j: y_{j,c}=1}(\text{softmax}(\bm {u}_{j})-\bm {y}_{j}),
\end{aligned}\label{lpl_max}
\end{equation}
where $\alpha$ is the hyper-parameter.
In the minimization part, $\bm{\tilde \delta} _{c}$ is updated by 

\begin{equation}
\begin{aligned}
\bm{\tilde{ \delta }}_{c} = -\frac{\alpha }{N_{c}} \sum_{j: y_{j,c}=1}(\text{softmax}(\bm {u}_{j})-\bm{y}_{j}).
\end{aligned}\label{lpl_min}
\end{equation}
The PGD-like optimization in Algorithm \ref{alg:2} contains two hyper-parameters, namely, step size and \#steps. Let $\alpha$ be the step size, and $K_{c}$ be the number of steps(\#steps) for category $c$. On the balanced classification, the $\alpha$  is searched in \{0.01, 0.02, 0.03\}. With step size, the PGD-like optimization is detailed in Algorithm \ref{alg:2}.

\begin{algorithm}[tb]
\caption{PGD-like Optimization} \label{alg:2}
\textbf{Input}: The logit vectors ($\bm u_{i}$) for the $c$th category in the current mini-batch, $\epsilon_c$, and $\alpha$.
\begin{algorithmic}[1] 
\STATE Let $\bm {u}^{0}_{i} = \bm u_{i}$ for the input vectors;
\STATE Calculate $K_{c}$ by $\lfloor \frac{\epsilon _{c}}{\alpha} \rfloor$;
\FOR{ $k$ = 1 to $K_{c}$}
\STATE  Calculate $\frac{\partial l(\text{softmax}(\bm {u}_{i}^k+\bm {\tilde{\delta}}_{c}), \bm y_i)}{\partial \bm {\tilde{\delta}}_{c}}\bigg|_{\boldsymbol{0}}=\text{softmax}(\bm {u}^{k}_{i})-\bm y_i$.
\STATE Calculate $\bm {\tilde{\delta}}_{c}^{k+1}$ according to Eq. (\ref{lpl_max}) for maximization and Eq. (\ref{lpl_min}) for minimization;
\STATE $\bm {u}^{k+1}_{i} := \bm {u}^{k}_{i} + \bm {\tilde{\delta}}_{c}^{k+1}$.
\ENDFOR
\end{algorithmic}\label{PGD-like}
\textbf{Output}: $\bm{\tilde\delta} _{c} = \bm u_{i}^{K_{c}} - \bm u_{i}$ 
\end{algorithm} 


\textbf{Bound calculation.}
The category with a relatively low/high performance should be more positively/negatively augmented; the category closer to the tail/head should be more positively/negatively augmented. We define
\begin{equation}
\begin{array}{l}
{\epsilon _c}{\rm{ = }}\epsilon {\rm{ + }}\Delta \epsilon \left| \tau  -  \bar q_c \right|, \\ {\rm{ or}} \
{\epsilon _c}{\rm{ = }}\left\{ {\begin{array}{*{20}{c}}
{\epsilon  + \Delta \epsilon \frac{{{{ \bar q}_c} }}{{{{\bar q}_1}}}}&{c \le \tau }\\
{\epsilon  + \Delta \epsilon \frac{{{{ \bar q}_C} }}{{{{\bar q}_c}}}}&{c > \tau }
\end{array}} \right.
\end{array}\label{finalbound_lpl}.
\end{equation} 
In Eq. (\ref{finalbound_lpl}), the larger the difference between the performance (${ \bar q}_c$) of the current category and the threshold $\tau$, or the larger the ratio ${ \bar q}_c/{\bar q}_1$ (and ${\bar q}_C/{\bar q}_c$), the larger the bound $\epsilon_c$. This notion is in accordance with our previous conjecture. When $\Delta \epsilon $ in Eq. (\ref{finalbound_lpl}) equals to zero, the bound is fixed. The algorithmic steps of our LPL for single-label learning are in Algorithm \ref{alg:3}.

\textbf{Comparative Analysis.} We compare the perturbations in ISDA and our LPL in terms of data augmentation.

In the ISDA's rationale, new samples are (virtually instead of really) generated based on the distribution of each category. Fig. \ref{fig_ISDA} shows the (virtually) generated samples by ISDA. In the right case, the positive augmentation for head category may further hurt the performance of the tail category. ISDA fails in the long-tail problem. Li et al.~\cite{li2021metasaug} leverage meta learning to adapt ISDA for the long-tail problem.

In contrast with the above-mentioned methods, our proposed LPL method conducts positive or negative augmentation according to the directions of loss maximization and minimization. According to our Corollaries 1-3, loss maximization will force the category to move close to the decision boundary (i.e., the category is positively augmented or virtual samples are generated for this category). By contrast, loss minimization will force the category to be far from the boundary (i.e., the category is negatively augmented or samples are virtually deleted for this category). Fig. \ref{fig_LPL} shows an illustrative example.

\subsection{Logit Perturbation Method (LPL) for Multi-label Learning}

\begin{figure*}[ht]
\centering
\includegraphics[width=0.9\textwidth, height=1.1in]{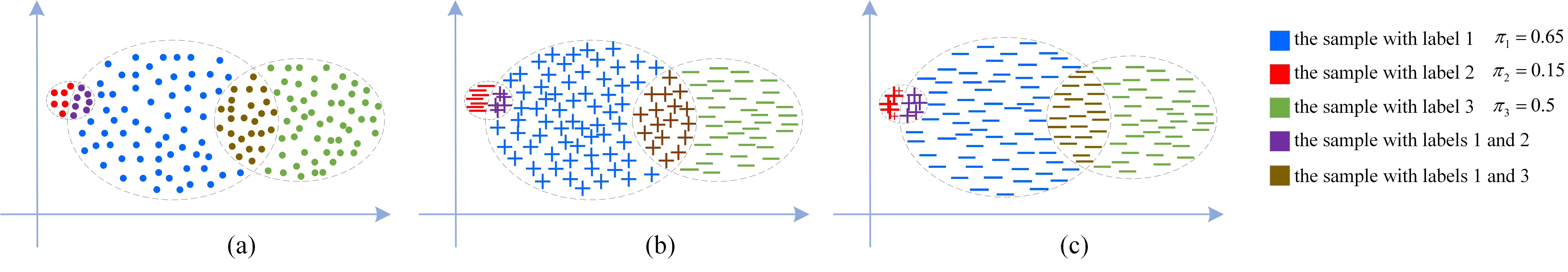}  \vspace{-0.15in}
\caption{An illustrative example for the variance imbalance and class imbalance in multi-label learning. ``$+$" means the positive samples.  ``$-$" means the negative samples. (a) shows a multi-label learning task ($C=3$). Different colors mean those samples with one label or more labels. (b) shows the case of variance imbalance. (c) shows the case of class imbalance.
}
\label{fig_multi_label_example}
\end{figure*}

Multi-label learning is usually decomposed into $C$ binary learning tasks. Compared with single-label learning, both variance imbalance and class imbalance usually exist in each of the $C$ tasks, simultaneously. First, variance imbalance exists in each of the $C$ tasks. The reason lies in that negative samples in each of the $C$ tasks actually come from the remaining $C-1$ classes, whereas positive samples in each task come from only one class. Naturally, the variance of the negative samples will be larger than that of the positive samples as shown in Fig. \ref{fig_multi_label_example} (b). Theoretically, the negative samples require the first type of logit perturbation and the positive samples require the second type of logit perturbation. Second, class imbalance may exist in each of the $C$ tasks as shown in Fig. \ref{fig_multi_label_example} (c). However, the class imbalance degrees for tasks in which the positive samples are from the tail categories are larger than those for tasks in which the positive samples are from the head categories. Therefore, according to Corollaries 1 and 2, the negative samples require the second type of logit perturbation, and the positive samples require the first type of logit perturbation (especially for the tasks when the positive samples belong to tail categories). 

Obviously, there is contradiction for the two cases discussed above. To deal with variance imbalance, the negative samples should perform the first type of logit perturbation. Meanwhile, to deal with class imbalance, the negative samples should perform the second type of logit perturbation. Corollary 3 demonstrates that the perturbation type is dependent of the primary challenge on the class or variance imbalances. Consequently, we extend Eq. (\ref{longtail_new_loss}) into the following form for multi-label learning.

\vspace{-0.2in}
 \begin{equation}
\begin{aligned}
\mathcal{L}&=\frac{1}{C \times N}\sum_{(\bm x_i,\bm y_i)}\sum_{c = 1}^{C}\mathbb{S}(c - \tau )\times
\\&\{ \max_{| {{{\tilde \delta }_{{c}}}} | \le \epsilon_c} y_{i,c}\log(1+e^{- u_{i,c}+{{{\tilde \delta }_{{c}}}}})\times \mathbb{S}(c - \tau )
\\&+\min_{| {{{\tilde \delta }_{{c}}}} | \le \epsilon_c}(1- y_{i,c})\log(1+e^{ u_{i,c}-{{{\tilde \delta }_{{c}}}}})\times \mathbb{S}(c - \tau )\},
\end{aligned}\label{multi_label_new_loss}
\end{equation} 
where ${{{\tilde \delta }_{{c}}}}$ is a scalar, and $\tau$ is a hyperparameter (threshold) for the category split. This new loss can effectively tune the cooperation of the two types of logit perturbation by setting an appropriate value of $\tau$. There are three typical settings for $\tau$ as shown as follows:
\begin{itemize}
\item If $\tau$ is set as zero, $\mathbb{S}(c - \tau ) \equiv 1$. Eq.~(\ref{multi_label_new_loss}) becomes
 \begin{equation}
\begin{aligned}
\mathcal{L}&=\frac{1}{C \times N}\sum_{(\bm x_i,\bm y_i)}\sum_{c = 1}^{C}
\{ \max_{| {{{\tilde \delta }_{{c}}}} | \le \epsilon_c} y_{i,c}\log(1+e^{- u_{i,c}+{{{\tilde \delta }_{{c}}}}})
\\&+\min_{| {{{\tilde \delta }_{{c}}}} | \le \epsilon_c}(1- y_{i,c})\log(1+e^{ u_{i,c}-{{{\tilde \delta }_{{c}}}}})\}.
\end{aligned}\label{multi_label_new_loss_1}
\end{equation} 
In this situation, the positive samples of all the $C$ tasks perform the first type of logit perturbation, indicating that class imbalance is the primary concern in all tasks. 

\item If $\tau$ is set as $C+1$, then $\mathbb{S}(c - \tau ) \equiv -1$. Eq.~(\ref{multi_label_new_loss}) becomes
 \begin{equation}
\begin{aligned}
\mathcal{L}&=\frac{1}{C \times N}\sum_{(\bm x_i,\bm y_i)}\sum_{c = 1}^{C}
\{ \min_{| {{{\tilde \delta }_{{c}}}} | \le \epsilon_c} y_{i,c}\log(1+e^{- u_{i,c}+{{{\tilde \delta }_{{c}}}}})
\\&+\max_{| {{{\tilde \delta }_{{c}}}} | \le \epsilon_c}(1- y_{i,c})\log(1+e^{ u_{i,c}-{{{\tilde \delta }_{{c}}}}})\},
\end{aligned}\label{multi_label_new_loss_2}
\end{equation} 
In this situation, the negative samples of all the $C$ binary tasks perform the second type of logit perturbation, indicating that variance imbalance is the primary concern in all tasks.

\item If $1 < \tau < C$, then $\mathbb{S}(c - \tau ) \equiv 1$ when $c > \tau$ and $\mathbb{S}(c - \tau ) \equiv -1$ when $c < \tau$. When $c<\tau$, the optimization for the $c$th binary task becomes
 \begin{equation}
\begin{aligned}
\mathcal{L}_c&=\frac{1}{ N}\sum_{(\bm x_i,\bm y_i)}\{ \min_{| {{{\tilde \delta }_{{c}}}} | \le \epsilon_c} y_{i,c}\log(1+e^{- u_{i,c}+{{{\tilde \delta }_{{c}}}}})
\\&+\max_{| {{{\tilde \delta }_{{c}}}} | \le \epsilon_c}(1- y_{i,c})\log(1+e^{ u_{i,c}-{{{\tilde \delta }_{{c}}}}})\},
\end{aligned}\label{multi_label_variance_imbalance}
\end{equation} 
which indicates that the positive samples perform the second type of logit perturbation and the negative samples perform the first type of logit perturbation. This is reasonable because the $c$th class belongs to the head categories and thus variance imbalance rather than the class imbalance is the primary concern. When $c>\tau$, the optimization for the $c$th binary task becomes
 \begin{equation}
\begin{aligned}
\mathcal{L}_c&=\frac{1}{ N}\sum_{(\bm x_i,\bm y_i)}\{ \max_{| {{{\tilde \delta }_{{c}}}} | \le \epsilon_c} y_{i,c}\log(1+e^{-  u_{i,c}+{{{\tilde \delta }_{{c}}}}})
\\&+\min_{| {{{\tilde \delta }_{{c}}}} | \le \epsilon_c}(1-  y_{i,c})\log(1+e^{ u_{i,c}-{{{\tilde \delta }_{{c}}}}})\},
\end{aligned}\label{multi_label_class_imbalance}
\end{equation} 
which presents that the positive samples perform the first type of logit perturbation and the negative samples perform the second type of logit perturbation. This is reasonable because the $c$th class belongs to the tail categories, and class imbalance rather than the variance imbalance becomes the primary concern in learning. 
\end{itemize}

The third setting is adopted in our experiments. Similarly, we can perform PGD-like maximization and minimization as Algorithm \ref{PGD-like}. According to Eq. (\ref{multi_label_new_loss}), for positive samples, the derivative of the loss with respect to ${{{\tilde \delta }_{{c}}}}$ is as follows.
\begin{equation}
    \begin{aligned}
    &\frac{\partial \log(1+e^{- u_{i,c}+{{{\tilde \delta }_{{c}}}}})}{\partial {{{\tilde \delta }_{{c}}}}}\bigg|_{\boldsymbol{0}} =1-\text{sigmoid}(- u_{i,c}).
    \end{aligned}\label{pos_deritive_multi_label}
\end{equation}
For negative sample, the derivative of the loss with respect to ${{{\tilde \delta }_{{c}}}}$ is as follows.
\begin{equation}
    \begin{aligned}
    \frac{\partial \log(1+e^{ u_{i,c}-{{{\tilde \delta }_{{c}}}}})}{\partial {{{\tilde \delta }_{c}}}}\bigg|_{\boldsymbol{0}}=\text{sigmoid}( u_{i,c})-1.
    \end{aligned}\label{neg_deritive_multi_label}
\end{equation}
According to Eq. (\ref{multi_label_new_loss}), we use Eqs. (\ref{pos_deritive_multi_label}) and (\ref{neg_deritive_multi_label}) to calculate ${{{\tilde \delta }_{c}}}$ as follows.
\begin{equation}
    \begin{aligned}
    &{\tilde \delta }_{c}=\frac{\alpha}{C\times N}\sum_{i=1}^{N}\{ y_{i,c}(\text{sigmoid}(- u_{i,c})-1)\\&+(1- y_{i,c})(\text{sigmoid}( u_{i,c})-1)\}\times \mathbb{S}(c - \tau )\}.\label{delta_update}
    \end{aligned}
\end{equation}
where $\alpha$ is the step size. Since each image is treated as $C$ binary classification tasks, we can further simplify Eq. (\ref{delta_update}).
Positive and negative samples for each of $C$ tasks share the same ${{{\tilde \delta }_{c}}}$ for the class $c$. Obviously, according to Eqs. (\ref{pos_deritive_multi_label}) and (\ref{neg_deritive_multi_label}), $1-\text{sigmoid}(- u_{i,c}) \ge 0$ and $\text{sigmoid}( u_{i,c})-1\le 0$ holds. The term ${\tilde \delta }_{{c}}$ is a scalar. Therefore, when the perturbation bound ${\epsilon _c}$ is given by Eq. (\ref{finalbound_lpl}), we can easily obtain ${\tilde \delta }_{{c}} = \epsilon_c$. Then the logit perturbation for multi-label learning can be easily calculated.

The algorithmic steps of our LPL for multi-label learning are also in Algorithm \ref{alg:3}.

\begin{algorithm}[tb]
\caption{Learning to Perturb Logits (LPL)} \label{alg:3}
\label{alg:algorithm}
\textbf{Input}: $\displaystyle S$, $\tau$, max iteration $T$, hyper-parameters for PGD-like optimization, and other conventional training hyper-parameters.\text{ }   
\begin{algorithmic}[1] 
\STATE Randomly initialize $\bm W$. \\
\FOR{ $t$ = 1 to $T$}
\STATE Sample a mini-batch from $\displaystyle S$.
\STATE Update $\tau$ if it is not fixed (e.g., ${{\text{mean}({\bar q_{c}})}}$ is used) and split the category set.
\STATE Compute ${\epsilon _c}$ for each category using Eq. (\ref{finalbound_lpl}) if varied bounds are used.
\STATE Infer $\bm{\tilde \delta}_c$ for each category using a PGD-like optimization method for Eq. (\ref{special_loss1}) in balanced classification, Eq. (\ref{longtail_new_loss}) in long-tail classification, or Eq. (\ref{multi_label_new_loss}) in multi-label classification.
\STATE   Update the logits for each sample and the loss.
\STATE Update $\bm W$ with SGD.
\ENDFOR
\end{algorithmic}
\textbf{Output}: $\bm W$
\end{algorithm}

\begin{table}[tb]

\caption{Mean values and standard deviations of the test Top-1 errors for all the involved methods on CIFAR10.}\label{tab:banlance_cifar10}
\centering
\vspace{-0.08in}
\begin{tabular}{|p{2.18cm}||c||c|}\hline
    Method &  Wide-ResNet-28-10 & ResNet-110  \\\hline
    Basic & 3.82 ± 0.15\% & 6.76 ± 0.34\% \\
    Large Margin & 3.69 ± 0.10\% & 6.46 ± 0.20\% \\
    Disturb Label & 3.91 ± 0.10\% & 6.61 ± 0.04\% \\
    Focal Loss & 3.62 ± 0.07\% & 6.68 ± 0.22\% \\
    Center Loss & 3.76 ± 0.05\% & 6.38 ± 0.20\% \\
    Lq Loss & 3.78 ± 0.08\% & 6.69 ± 0.07\% \\
    CGAN & 3.84 ± 0.07\% & 6.56 ± 0.14\% \\
    ACGAN & 3.81 ± 0.11\% & 6.32 ± 0.12\% \\
    infoGAN & 3.81 ± 0.05\% & 6.59 ± 0.12\% \\
    ISDA & 3.58 ± 0.15\% & 6.33 ± 0.19\% \\
    ISDA+DropOut & 3.58 ± 0.15\% & 5.98 ± 0.20\% \\\hline
    \multicolumn{1}{|m{2.18cm}||}{LPL (mean+ fixed $\epsilon_c$)}  & 3.39 ± 0.04\% & 5.83 ± 0.21\% \\
    \multicolumn{1}{|m{2.18cm}||}{LPL ({mean+ varied $\epsilon_c$})} & \textbf{3.37 ±  0.04}\% & \textbf{5.72 ± 0.05}\% \\\hline
\end{tabular}
\end{table}

\begin{table}[tb]
\caption{Mean values and standard deviations of the test Top-1 errors for all the involved methods on CIFAR100.}\label{tab:banlance_cifar100}
\centering
\vspace{-0.08in}
\begin{tabular}{|p{2.18cm}||c||c|}\hline
    Method & Wide-ResNet-28-10 & ResNet-110  \\\hline
    Basic & 18.53 ± 0.07\% & 28.67 ± 0.44\% \\
    Large Margin & 18.48 ± 0.05\% & 28.00 ± 0.09\% \\
    Disturb Label & 18.56 ± 0.22\% & 28.46 ± 0.32\% \\
    Focal Loss & 18.22 ± 0.08\% & 28.28 ± 0.32\% \\
    Center Loss & 18.50 ± 0.25\% & 27.85 ± 0.10\% \\
    Lq Loss & 18.43 ± 0.37\% & 28.78 ± 0.35\% \\
    CGAN & 18.79 ± 0.08\% & 28.25 ± 0.36\% \\
    ACGAN & 18.54 ± 0.05\% & 28.48 ± 0.44\% \\
    infoGAN & 18.44 ± 0.10\% & 27.64 ± 0.14\% \\
    ISDA & 17.98 ± 0.15\% & 27.57 ± 0.46\% \\
    ISDA+DropOut & 17.98 ± 0.15\% & 26.35 ± 0.30\% \\\hline
    \multicolumn{1}{|m{2.18cm}||}{LPL (mean+ fixed $\epsilon_c$)} & 18.19 ± 0.07\% & 26.09 ± 0.16\% \\
    \multicolumn{1}{|m{2.18cm}||}{LPL ({mean+ varied $\epsilon_c$})} & \textbf{17.61 ± 0.30}\% & \textbf{25.87 ± 0.07\%} \\\hline
\end{tabular} 

\label{table_CIFAR100_balance}
\end{table}

\section{Experiments}
Our proposed LPL is first evaluated on data augmentation, long-tail classification and multi-label classification tasks. The properties of LPL are then analyzed with more experiments. A Linux platform with four RTX3090 graphics cards is used, and each graphics card has a capacity of 24 GB.

\subsection{Experiments on Data Augmentation}

\textbf{Datasets and competing methods}. In this subsection, two benchmark image classification data sets, namely, CIFAR10 \cite{krizhevsky2009learning} and CIFAR100 \cite{krizhevsky2009learning}, are used. Both data consist of 32$\times$32 natural images in 10 classes for CIFAR10 and 100 classes for CIFAR100. There are 50,000 images for training and 10,000 images for testing. The training and testing configurations used in~\cite{wang2019implicit} are followed. Several classical and state-of-the-art robust loss functions and (semantic) data augmentation methods are compared: Large-margin loss~\cite{liu2016large}, Disturb label~\cite{xie2016disturblabel}, Focal Loss~\cite{lin2017focal}, Center loss~\cite{wen2016discriminative}, Lq loss~\cite{zhang2018generalized},
CGAN~\cite{mirza2014conditional}, ACGAN~\cite{odena2017conditional}, infoGAN~\cite{chen2016infogan}, ISDA, and ISDA + Dropout.

 The Wide-ResNet-28~\cite{zagoruyko2016wide} and ResNet-110~\cite{he2016deep} are used as the base neural networks. Considering that the training/testing configuration is fixed for both sets, the results of the above competing methods reported in the ISDA paper~\cite{wang2019implicit} are directly presented (some are from their original papers). The training settings for the above base neural networks also follow the instructions of ISDA paper and its released codes. Our methods have two variants.
\begin{itemize}
\item LPL (mean+fixed bound). In this version, the optimization in Eq. (\ref{special_loss1}) is used. Mean denotes that the threshold is $\text{mean}({\bar q_{c}})$. Fixed bound means that the value of $\epsilon_c$ is fixed and identical for all categories during optimization. It is searched in \{0.1, 0.2, 0.3, 0.4\}.
\item LPL (mean+varied bound). In this version, the optimization in Eq. (\ref{special_loss1}) is used. Theoretically, varied bound means that the value of $\epsilon_c$ is varied according to Eq. (\ref{finalbound_lpl}). However, the varied bounds in the same batch make the implementation more difficult and increase the training complexity. In our implementation, we choose to set a varied number of updating steps for each category in our PGD-like optimization. The value of $\Delta \epsilon$ is searched in \{0.1, 0.2\}.
\end{itemize}

The Top-1 error is used as the evaluation metric. The performances of the base neural networks with the standard cross-entropy loss are re-run before running our methods to conduct a fair comparison. Almost identical results are obtained compared with the published results in the ISDA paper. 

\textbf{Results}. Tables \ref{tab:banlance_cifar10} and \ref{tab:banlance_cifar100} present the results of all competing methods on the CIFAR10 and CIFAR100, respectively. Our LPL method (two versions) achieves the best performance almost under both the two base neural networks. ISDA achieves the second-best performance. Only in the case of Wide-ResNet-28-10 on CIFAR100, LPL (mean+fixed $\epsilon_c$) is inferior to ISDA. However, the former still achieves the fourth lowest error.


The results of LPL with varied bounds are better than those of LPL with fixed bounds. This comparison indicates the rationality of our motivation that the category with relatively low (high) performance should be more positively (negatively) augmented. In the final part of this section, more analyses will be conducted to compare ISDA and our method. Naturally, the varied threshold will further improve the performances.

\subsection{Experiments on Long-tail Classification}

\textbf{Datasets and competing methods}. In comparison with the conference version of the paper, we supplement the experiments with real-world data sets. In the synthetic data set experiment, the long-tail versions of CIFAR10 and CIFAR100 compiled by Cui et al.~\cite{cui2019class} are used and called CIFAR10-LT and CIFAR100-LT, respectively,  The training and testing configurations used in~\cite{menon2020long} are followed. In the real-world data set experiment, large-scale data sets iNaturalist 2017 (iNat2017)\cite{van2018inaturalist} and iNaturalist 2018 (iNat2018)\cite{iNaturalist2018} with extremely imbalanced class distributions are used. iNat2017 includes 579,184
training images in 5,089 classes with an imbalance factor
of 3919/9, while iNat2018 is composed of
435,713 images from 8,142 classes with an imbalance factor of 1000/2. Several classical and state-of-the-art robust loss functions and semantic data augmentation methods are compared: Class-balanced CE loss~\cite{wang2019implicit}, Class-balanced fine-tuning~\cite{Cui4190}, Meta-weight net~\cite{shu2019meta}, Focal loss~\cite{lin2017focal}, Class-balanced focal loss~\cite{cui2019class}, LDAM~\cite{cao2019learning}, LDAM-DAR~\cite{cao2019learning}, ISDA, and LA.

In the synthetic data set experiment, Menon et al.~\cite{menon2020long} released the training data when the imbalance ratio (i.e., $\pi_1/\pi_{100}$) is 100:1; hence, their data and reported results for the above competing methods are directly presented. When the ratio is 10:1, the results  of ISDA+Dropout and LA are obtained by running their released codes. The results of the rest methods are from the study conducted by Li et al.~\cite{li2021metasaug}. The hyper-parameter $\lambda$ in LA is searched in \{0.5, 1, 1.5, 2, 2.5\} according to the suggestion in ~\cite{menon2020long}. Similar to the experiments in~\cite{menon2020long}, ResNet-32~\cite{he2016deep} is used as the base network. The results of ISDA, LA, and LPL are the average of five repeated runs.

In the real-world data set experiment, the results
of the above competing methods reported in \cite{menon2020long} are directly presented. The results of LA on iNat2018 are from the original paper~\cite{menon2020long}. The other results, such as ISDA+dropout and LA on iNat2017, are obtained by running their released codes. Likewise, the hyper-parameter $\lambda$ in LA is searched in \{0.5, 1, 1.5, 2, 2.5\}. Similar to the experiments in~\cite{wu2020dist}, ResNet-50~\cite{he2016deep} is used as the base network.  All results are the average of five repeated runs. 

Our methods have two variants: LPL (varied threshold + fixed bound) and LPL (varied threshold + varied bound). The threshold $\tau$ is searched in \{0.4$C$, 0.5$C$, 0.6$C$\}. In the fixed bound version, the value of $\Delta \varepsilon $ is set to 0, and $\epsilon$ is searched in \{1.5, 2.5, 5\}.  In the varied bound version, the value of $\epsilon$ is set to 0, and $\Delta \varepsilon $ is searched in \{1.0, 2.0, 3.0\}.
Only one meta-based method, Meta-weight net, is involved, because we mainly aim to compare methods that only modify the training loss. In addition, meta-based methods require an auxiliary high-quality validation set~\cite{li2021metasaug}. Other methods, such as BBN~\cite{zhou2020bbn}, which focus on the new network structure are also not included in the comparisons.

\begin{table}[tb]
\caption{Test Top-1 errors on CIFAR100-LT (ResNet-32).}\label{tab:longtail_cifar100}
\centering
\vspace{-0.08in}
\begin{tabular}{|p{3.9cm}||c||c|}\hline
    Ratio & 100:1 & 10:1 \\\hline
    Class-balanced CE loss  & 61.23\% & 42.43\% \\
    Class-balanced fine-tuning  & 58.50\% & 42.43\% \\
    Meta-weight net  & 58.39\% & 41.09\% \\
    Focal Loss & 61.59\% & 44.22\% \\
    Class-balanced focal loss  & 60.40\% & 42.01\% \\
    LDAM  & 59.40\% & 42.71\% \\
    LDAM-DRW  & 57.11\% & 41.22\% \\
    ISDA + Dropout & 62.60\% & 44.49\% \\
    LA & 56.11\% & 41.66\% \\\hline
    LPL (varied $\tau$ + fixed $\epsilon_c$) & 58.03\% & 41.86\% \\
    LPL (varied $\tau$ + varied $\epsilon_c$) & \textbf{55.75}\% & \textbf{39.03\%} \\\hline
\end{tabular} 
\end{table}

\begin{table}[!t]
\caption{Test Top-1 errors on CIFAR10-LT (ResNet-32).}\label{tab:longtail_cifar10}
\centering
\vspace{-0.08in}
\begin{tabular}{|p{3.9cm}||c||c|}\hline
    Ratio & 100:1 & 10:1 \\\hline
    Class-balanced CE loss & 27.32\% & 13.10\% \\
    Class-balanced fine-tuning  & 28.66\% & 16.83\% \\
    Meta-weight net & 26.43\% & 12.45\% \\
    Focal Loss & 29.62\% & 13.34\% \\
    Class-balanced focal loss  & 25.43\% & 12.52\% \\
    LDAM  & 26.45\% & 12.68\% \\
    LDAM-DRW  & 25.88\% & 11.63\% \\
    ISDA + Dropout & 26.45\% & 12.98\% \\
    LA & 22.33\% & 11.07\% \\\hline
    LPL (varied $\tau$ + fixed $\epsilon_c$) & 23.97\% & 11.09\% \\
    LPL (varied $\tau$ + varied $\epsilon_c$) & \textbf{22.05}\% & \textbf{10.59\%} \\\hline
\end{tabular}

\end{table}

\begin{table*}[htb]
\caption{ Results of mAP by our methods and other comparing
approaches on VOC-MLT and COCO-MLT. }\label{tab:multi-label_cifar100}
\centering
\vspace{-0.08in}
\resizebox{.8\textwidth}{!}{
\begin{tabular}{|l||c||c||c||c||c||c||c||c|}\hline
    Datasets & \multicolumn{4}{c||}{VOC-MLT} &  \multicolumn{4}{c|}{COCO-MLT}   \\\hline
    Method & total & head & medium & tail & total & head & medium & tail \\\hline
    ERM   & 70.86\% & 68.91\% & 80.20\% & 65.31\% & 41.27\% & 48.48\% & 49.06\% & 24.25\% \\
    RW   & 74.70\% & 67.58\% & 82.81\% & 73.96\% & 42.27\% & 48.62\% & 45.80\% & 32.02\% \\
    Focal Loss   & 73.88\% & 69.41\% & 81.43\% & 71.56\% & 49.46\% & 49.80\% & 54.77\% & 42.14\% \\
    RS   & 75.38\% & 70.95\% & 82.94\% & 73.05\% & 46.97\% & 47.58\% & 50.55\% & 41.70\% \\
    RS-Focal   & 76.45\% & 72.05\% & 83.42\% & 74.52\% & 51.14\% & 48.90\% & 54.79\% & 48.30\% \\
    ML-GCN   & 68.92\% & 70.14\% & 76.41\% & 62.39\% & 44.24\% & 44.04\% & 48.36\% & 38.96\% \\
    OLTR   & 71.02\% & 70.31\% & 79.80\% & 64.95\% & 45.83\% & 47.45\% & 50.63\% & 38.05\% \\
    LDAM   & 70.73\% & 68.73\% & 80.38\% & 69.09\% & 40.53\% & 48.77\% & 48.38\% & 22.92\% \\
    CB-Focal   & 75.24\% & 70.30\% & 83.53\% & 72.74\% & 49.06\% & 47.91\% & 53.01\% & 44.85\% \\
    R-BCE   & 76.34\% & 71.40\% & 82.76\% & 75.22\% & 49.43\% & 48.77\% & 53.00\% & 45.33\% \\
    R-BCE-Focal   & 77.39\% & 72.44\% & 83.16\% & 76.77\% & 52.75\% & 50.20\% & 56.52\% & 50.02\% \\
    
    R-BCE+NTR   & 78.65\% & 73.16\% & 84.11\% & 78.66\% & 52.53\% & 50.25\% & 56.33\% & 49.54\% \\
    R-BCE-Focal+NTR  & 78.94\% & 73.22\% & 84.18\% & 79.30\% & 53.55\% & 51.13\% & 57.05\% & 51.06\% \\
    R-BCE+LC  & 78.08\%  & 73.10\% & 83.49\% & 77.75\% & 53.68\% & 50.58\% & 57.10\% & 51.90\% \\
    R-BCE-Focal+LC  & 78.66\% & 72.74\% & 83.45\% & 79.52\% & 53.94\% & 50.99\% & 57.47\% & 51.88\% \\\hline
    R-BCE+LPL (varied $\tau$ + fixed $\epsilon_c$) & 78.64\%  & 73.00\% & 82.81\% & 79.74\% & 53.97\% & 50.23\% & 57.36\% & 52.79\% \\
    R-BCE+LPL (varied $\tau$ + varied $\epsilon_c$)  & 79.02\%  & 72.39\% & 82.14\% & 81.64\% & 54.35\% & 51.48\% & 57.72\% & 52.42\% \\
    R-BCE-Focal+LPL (varied $\tau$ + fixed $\epsilon_c$)  & 79.17\%  & 73.33\% & 83.56\% & 80.27\% & 54.37\% & 51.14\% & 57.68\% & 52.85\% \\
    R-BCE-Focal+LPL (varied $\tau$ + varied $\epsilon_c$)  & \textbf{79.57}\%  & 73.47\% & 83.95\% & 80.87\% & \textbf{54.76}\% & 50.78\% & 58.12\% & 53.81\% \\\hline
\end{tabular}}
\end{table*}

\begin{table}[h]
\caption{Test Top-1 errors on real-world datasets (ResNet-50). }\label{tab:longtail_inat}
\centering
\vspace{-0.08in}
\begin{tabular}{|p{3.8cm}||c||c|}\hline
    Method  & iNat2017 & iNat2018  \\\hline
    Class-balanced CE loss  & 42.02\% & 33.57\% \\
    Class-balanced fine-tuning & -- &  -- \\
    Meta-weight net  & -- & -- \\
    Focal Loss& --  & -- \\
    Class-balanced focal loss & 41.92\% & 38.88\% \\
    LDAM    & 39.15\% & 34.13\% \\
    LDAM-DRW    &37.84\% & 32.12\%  \\
    ISDA + Dropout    & 43.37\% & 39.92\% \\
    LA    & 36.75\% & 31.56\% \\\hline
    LPL (varied $\tau$ + fixed $\epsilon_c$)    & 38.47\% & 32.06\% \\
    LPL (varied $\tau$ + varied $\epsilon_c$)  & \textbf{35.86}\% & \textbf{30.59}\% \\\hline
\end{tabular}
\end{table}

\begin{table}[b]
\caption{The error reduction of LPL (varied $\tau$+varied $\epsilon$) over LA on the two data sets.}\label{tab:LPL_vs_LA}
\centering
\vspace{-0.08in}
\begin{tabular}{|c||p{1.1cm}||p{1.1cm}||p{1.1cm}||p{1.1cm}|}\hline
    Ratio & \multicolumn{2}{c||}{100:1} &  \multicolumn{2}{c|}{10:1}   \\\hline
    LA & 56.11\% & 22.33\% & 41.66\% & 11.07\% \\\hline
    LPL & 55.75\% (-0.36\%) & 22.05\% (-0.28\%) & 39.03\% (-2.63\%) & 10.59\% (-0.48\%) \\\hline
\end{tabular}

\end{table}


\textbf{Results}. The Top-1 error is also used. Table \ref{tab:longtail_cifar100} shows the results of all the methods on the CIFAR100-LT data. On the ratios 100:1 and 10:1, LPL (varied $\tau$ + varied $\epsilon_c$) yields the lowest Top-1 errors. It exceeds the best competing method LA by 0.36\% and 2.63\% on the ratios 100:1 and 10:1, respectively. Table \ref{tab:longtail_cifar10} shows the results of all the methods on the CIFAR10-LT data. LPL (varied $\tau$ + varied $\epsilon_c$) still obtains the lowest Top-1 errors on both ratios. Table \ref{tab:longtail_inat} shows the results of all the methods on the iNat2017 and iNat2018. For real-world long-tail datasets, it still exceeds LA 0.89\% and 0.97\%, respectively. On all the comparisons, the semantic augmentation method ISDA obtains poor results. On CIFAR100-LT, ISDA achieves the worst performances on both ratios. This result is expected because ISDA aims to positively augment all categories equally and does not favor tail categories, which may lead to tail categories suffering from this positive augmentation. Nevertheless, ISDA has a better performance on CIFAR10-LT than on CIFAR100-LT. In Fig. \ref{relative_loss_variations_single} (b), the loss increments of tail categories are larger than those of the head ones. That is, larger augmentations are exerted on tail categories.

We listed the Top-1 errors of LA and LPL (varied $\tau$ + varied $\epsilon_c$) on Table \ref{tab:LPL_vs_LA} to better present the comparison. When the ratio is smaller, the improvements (error reductions) are relatively larger. This result is reasonable because when the ratio becomes small, the effectiveness of LA will be subsequently weakened. When the imbalance ratio is one, indicating that there is no imbalance, LA will lose effect; however, our LPL can still augment the training data effectively.

\begin{table*}[htb]
\caption{Number of parameters and test Top-1 errors of ISDA and LPL with different base networks.}
\label{tab:other_lpl}
\centering
\vspace{-0.08in}
\resizebox{.7\textwidth}{!}{
\begin{tabular}{|l||c||c||c|}\hline
    Method & \#Params & CIFAR10 & CIFAR100 \\\hline
    ResNet-32+ISDA & 0.5M & 7.09 ± 0.12\% & 30.27 ± 0.34\% \\
    ResNet-32+LPL (mean + fixed $\epsilon_c$) & 0.5M & 7.01 ± 0.16\% & 29.59 ± 0.27\% \\
    ResNet-32+LPL (mean + varied $\epsilon_c$) & 0.5M & \textbf{6.66 ± 0.09\%} & \textbf{28.53 ± 0.16\%} \\\hline
    SE-Resnet110+ISDA & 1.7M & 5.96 ± 0.21\% & 26.63 ± 0.21\% \\
    SE-Resnet110+LPL (mean + fixed $\epsilon_c$) & 1.7M & 5.87 ± 0.17\% & 26.12 ± 0.24\% \\
    SE-Resnet110+LPL (mean + varied $\epsilon_c$) & 1.7M & \textbf{5.39 ± 0.10\%}  & \textbf{25.70 ± 0.07\%} \\\hline
    Wide-ResNet-16-8+ISDA & 11.0M & 4.04 ± 0.29\% & 19.91 ± 0.21\% \\
    Wide-ResNet-16-8+LPL (mean + fixed $\epsilon_c$) & 11.0M & 3.97 ±  0.09\% & 19.87 ± 0.02\% \\
    Wide-ResNet-16-8+LPL (mean + varied $\epsilon_c$) & 11.0M & \textbf{3.93 ±  0.10\%} & \textbf{19.83 ± 0.09\%} \\\hline
\end{tabular} 
}
\end{table*}

\subsection{Experiments on Multi-label Classification}

\textbf{Datasets and competing methods}. In this part, the long-tail multi-label versions of VOC\cite{everingham2015pascal} and MS-COCO\cite{lin2014microsoft}  compiled by Wu et al.~\cite{wu2020dist} are used and called VOC-MLT and COCO-MLT, respectively. The training and test configurations used in~\cite{wu2020dist} are followed. The training set of VOC-MLT is sampled from train-val set of
VOC2012, containing 1142 images from 20 categories, with a maximum of 775 images per category and a minimum of 4 images per category. A toal of 4952 images from the VOC2007  test set are used for evaluation. COCO-MLT  is sampled from MS COCO-2017 dataset, containing 1909 images from 80 categories, with a maximum of 1128 images per category and a minimum of 6 images per category. 5000 images from the MS COCO-2017 test set are used for evaluation.

We mainly compare NTR and LC that perturb logit. The code of LC is not open sourced.
To keep the consistency of the experimental setup, we conduct both comparison experiments on the basis of R-BCE \cite{wu2020dist}. Several classical and state-of-the-art robust loss functions and multi-label methods are compared:
Empirical Risk Minimization (ERM), Re-Weighting (RW), Focal Loss~\cite{lin2017focal}, Re-Sampling (RS)~\cite{shen2016relay}, ML-GCN~\cite{chen2019multi}, OLTR~\cite{liu2019large}, LDAM~\cite{cao2019learning}, 
CB-Focal~\cite{cui2019class}, 
R-BCE ~\cite{wu2020dist}, 
R-BCE-Focal ~\cite{wu2020dist}, 
R-BCE + NTR ~\cite{wu2020dist}, 
R-BCE-Focal + NTR ~\cite{wu2020dist}, 
R-BCE + LC ~\cite{guo2021long}, 
R-BCE-Focal + LC~\cite{guo2021long}.

Wu et al.~\cite{wu2020dist} released the training data and code. Hence, their data and reported results for the above competing methods are directly presented. The experimental results of LC are reimplemented from the original paper's formula. Similar to the experiments in~\cite{wu2020dist}, ResNet-50~\cite{he2016deep} is used as the base network.

Our methods have two variants: LPL (varied threshold + fixed bound) and LPL (varied threshold + varied bound). The threshold $\tau$ is searched in \{0.4$C$, 0.5$C$, 0.6$C$\}. In the fixed bound version, the value of $\Delta \varepsilon $ is set to 0, and $\epsilon$ is searched in \{0.05, 0.1, 0.1\}.  In the varied bound version, the value of $\epsilon$ is set to 0, and $\Delta \varepsilon $ is searched in \{0.1, 0.2, 0.3\}. Other experimental setups such as training epochs and optimizer follow NTR.

\textbf{Results}. The evaluation metric mAP is used. Table \ref{tab:multi-label_cifar100} shows the results of all the methods on VOC-MLT and COCO-MLT. Our method achieves competitive or better results. R-BCE-Focal+LPL (varied $\tau$ + varied $\epsilon_c$) achieves the best results on VOC-MLT and COCO-MLT. R-BCE-Focal+LPL (varied $\tau$ + varied $\epsilon_c$) outperforms R-BCE-Focal + NTR by 0.63\% and 1.21\%, respectively, and outperforms R-BCE-Focal + LC by 0.91\% and 0.82\%, respectively. In the comparison experiment, R-BCE-Focal+LPL (varied $\tau$ + varied $\epsilon_c$) exceeds R-BCE-Focal by 2.18 \% on VOC-MLT and by 2.01 \% on COCO-MLT, respexctively. Similarly, when our method is added to the baseline R-BCE, our method can further improve the performance. The effectiveness of LPL is well proven.

\begin{table}[b]
\caption{Test Top-1 errors of three methods on two data sets.}\label{tab:table_LPL+LA}
\centering
\vspace{-0.08in}
\begin{tabular}{|c||c||c|}\hline
    Method & CIFAR10-LT100 & CIFAR100-LT100 \\\hline
    LA  & 22.33\% & 56.11\% \\
    LPL  & 22.05\% & 55.75\% \\
    LA+LPL  & \textbf{21.46\%} & \textbf{53.89\%} \\\hline
\end{tabular}
\end{table}

\begin{table}[b]
\caption{Results of mAP by our methods and other comparing
approaches on MS-COCO.}\label{tab:multi-label-big}
\centering
\vspace{-0.08in}
\begin{tabular}{|l||c|}\hline
    Method  & MS-COCO \\\hline
    R-BCE+NTR  & 83.7\% \\
    R-BCE+LC   & 84.5\% \\
    R-BCE+LPL(varied $\tau$ + varied $\epsilon_c$)  & \textbf{85.4\%} \\\hline

\end{tabular}\label{oriMLT}
\end{table}

\subsection{More Analysis for Our Method}

\textbf{Improvements on existing methods}. Our LPL method seeks the perturbation via an optimization scheme. In ISDA and LA, the perturbations are directly calculated rather than optimization. A natural question arises, that is, whether the perturbations in existing methods further improved via our method. Therefore, we propose a combination method with the following loss in imbalance image classification:
\begin{equation*}
\begin{aligned}
&\sum\limits_{{\rm{c}} \in {\bm {\mathcal{N}}_a}} {\sum\limits_{{\bm x_i} \in {\bm S_c}} {\mathop {\min }\limits_{\left\| {\bm {\tilde \delta} _{{c}}} \right\| \le \epsilon_c } l(\text{softmax} ({\bm u_i} + \lambda \log \boldsymbol{\pi} + \bm {\tilde \delta} _{{c}}),\bm y_i)} } \\
&+\sum\limits_{{\rm{c}} \in {\bm {\mathcal{P}}_a}} {\sum\limits_{{\bm x_i} \in {\bm S_c}} {\mathop {\max }\limits_{\left\| {\bm {\tilde \delta} _{{c}}} \right\| \le \epsilon_c } l(\text{softmax} ({\bm u_i} + \lambda \log \boldsymbol{\pi} + \bm {\tilde \delta} _{{c}}),{\bm y_i})} },
\end{aligned}
\end{equation*}
where $\log \boldsymbol{\pi} = [\log \pi_1, \cdots, \log \pi_C]$. When all $\epsilon_c$s are zero, the above-mentioned loss becomes the loss of LA; when $\lambda$ is zero, the above loss becomes our LPL (with fixed bound). We conducted experiments on CIFAR10-LT100 and CIFAR100-LT100. The results are shown in Table \ref{tab:table_LPL+LA}. ResNet-32 is used as the basic model. 
The value of $\lambda$ is searched in \{0.5, 1, 1.5, 2, 2.5\}.
The threshold $\tau$ is set as 4 and 40 on CIFAR10 and CIFAR100, respectively. Other parameters follow the setting in the previous experiments.

The combination method LA+LPL achieves the lowest errors on both comparisons, indicating that our LPL can further improve the performances of existing SOTA methods. ISDA can likewise be improved with the same manner.

\begin{figure}[htb]
\centering
\includegraphics[width=0.96\columnwidth, height=3.8in]{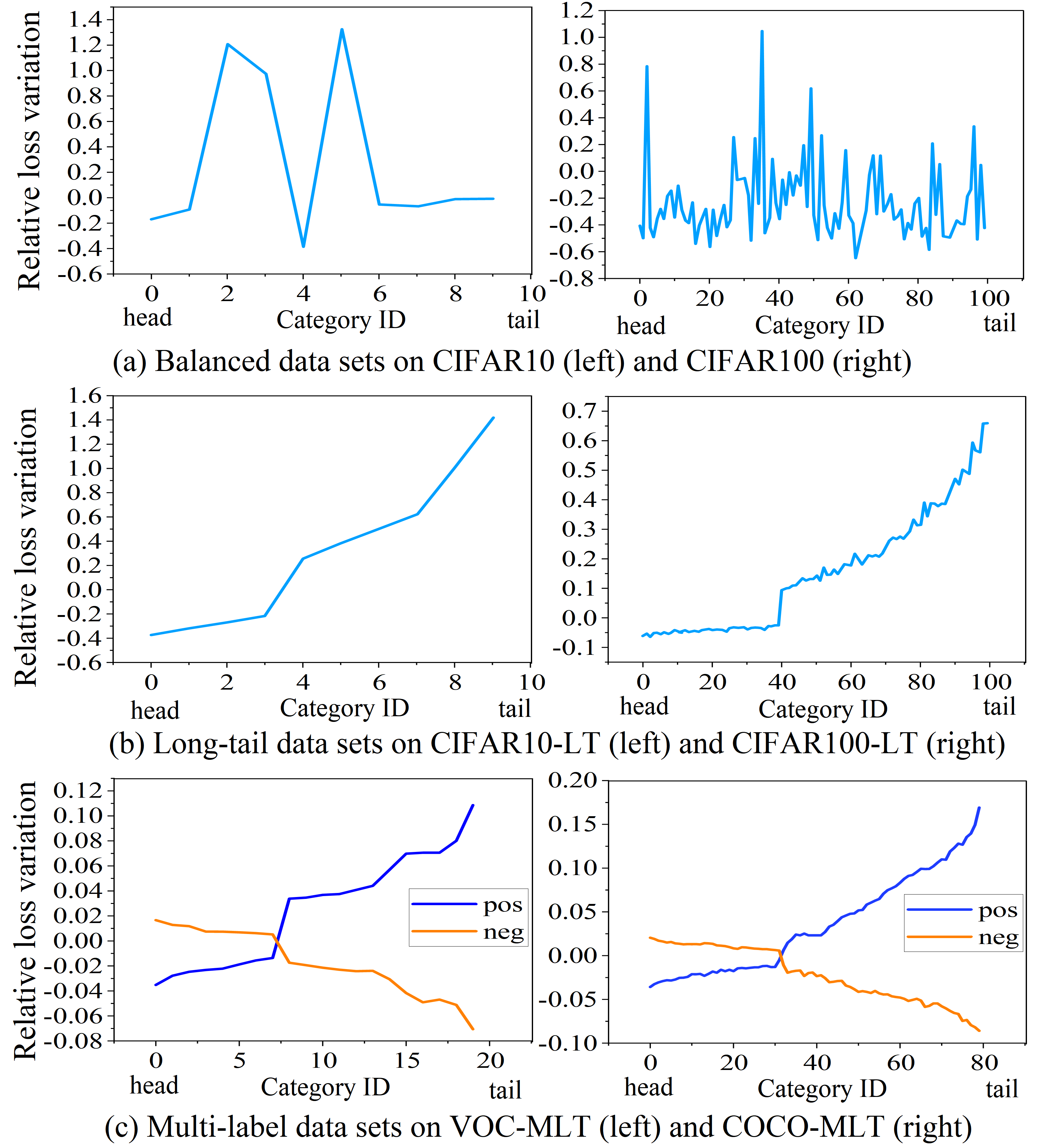}  \vspace{-0.1in}
\caption{Relative loss variations of our LPL on two balanced data sets, two long-tail data sets, and two multi-label data sets. ``pos" means the relative loss variations of positive samples. ``neg" means the relative loss variations of negative samples.}
\label{fig_loss_variation_lpl}
\end{figure}



\textbf{More comparisons with ISDA}. ISDA claims that it does not increase the number of parameters compared with the direct learning with the basic DNN models. Our method also does not increase the number of model parameters. The reason lies in that the perturbation terms are no longer used in the final prediction. 

Table \ref{tab:other_lpl} shows the comparisons between ISDA and LPL (two variants) on three additional base DNN models, namely, SE-ResNet110~\cite{hu2018squeeze}, Wide-ResNet-16~\cite{zagoruyko2016wide}, and ResNet-32.  The numbers of parameters are equal for ISDA and LPL. Nevertheless, the two variants of our method LPL outperform ISDA on both data sets under all the five base models.

\textbf{Loss variations of LPL during training}. For single-label classification, we plot the loss variations of LPL on two balanced and two long-tail data sets to assess whether our method LPL is in accordance with the two conjectures. The curves are shown in Fig. \ref{fig_loss_variation_lpl} (a) and (b). On the balanced data, the relative loss variations are similar to those of ISDA; on the long-tail data,
the losses of head categories are reduced, whereas those of tail ones are increased, which is similar to those of LA. For the multi-label classification, Fig. \ref{fig_loss_variation_lpl} (c) shows the results.
In comparison with NTR and LC, our method LPL focuses more on the tail categories according to the trends of relative loss reduction.

\textbf{Performances of LPL under different $\tau$ and $\epsilon_c$}. Both the threshold for category set split and the bound for augmentation extent are two important hyper-parameters in LPL. Based on our experiments, the following observations are obtained. On the balanced data sets, the results are relatively stable when the bound locates in [0.1, 0.5]; when the threshold is searched around the $\text{mean}({\bar q_{c}})$, the results are usually better. On the long-tail data sets, the results are relatively stable when the bound locates in [1.5, 5.0]. When the threshold is searched in \{0.4$C$, 0.5$C$, 0.6$C$\}, the results are usually good in our experiment. Long-tail problems require larger extent of data augmentation.

\textbf{More comparisons with NTR and LC}. We also compare our method with NTR and LC on the original multi-label dataset MS-COCO. MS-COCO contains 122,218 images with 80 different labels, which is divided to a training set with 82,081 images and a test set with 40,137 images. In this part, ResNet-110 is used as backbone network and the input size is 448$\times$448. Other setups follow Subsection C in Section IV. Table \ref{oriMLT} shows the results. The evaluation metric mAP is used. Again, our method achieves the competitive results. R-BCE+LPL(varied $\tau$ + varied $\epsilon_c$) exceeds R-BCE+NTR and R-BCE+LC 1.7 \% and 0.9 \% respectively.

\section{Conclusions}
This study investigates the class-level logit perturbation in deep learning. Two conjectures for the relationship between (logit perturbation-incurred) loss increment/decrement and positive/negative data augmentation are proposed. To support the two conjectures, theoretical investigation is performed in the presence of class imbalance and variance imbalance. On the basis of the two conjectures and our theoretical findings, new methodologies are introduced to learn to perturb logits (LPL) during DNN training for both single-label and multi-label learning tasks. Two key components of LPL, namely, category-set split and boundary calculation, are investigated. Extensive experiments on data augmentation (for balanced classification), long-tail classification, and multi-label classification are conducted. LPL achieves the best performances in both situations under different basic networks. Existing methods with logit perturbation (e.g. LA) can also be improved by using our method.

\bibliographystyle{IEEEtran} 
\bibliography{main.bib}

\section*{Appendix}
\subsection{Proof for Theorem 1}
\begin{proof}
Xu et al.~\cite{xu2021robust} proved that  $\boldsymbol{w}=\textbf{1}$ when the data distribution in Eq. (\ref{data_distribution}) is given (Lemma 1 in \cite{xu2021robust}). According to Lemma 1 in \cite{xu2021robust}, we can easily prove that when $P_{+}: P_{-}=1: \Gamma $ and $ \Gamma >1$, $\boldsymbol{w}=\textbf{1}$ holds. Thus, $f(\bm x)=\sum_{i=1}^{d}x_{i}+b$.
Then Eq. (\ref{optf_thm1}) can be written as follows.
\begin{equation}
   b^*=\arg\underset{b}{ \min } \operatorname{Pr}.(\mathbb{S}(\sum_{i=1}^{d}x_{i}+b+{{\tilde \delta }_{{c}}}^*) \neq y). \label{robf1}
\end{equation}
Now, we can calculate the optimal $b^*$ when the logit perturbation is used. Then, the optimal linear classifier is $f(\bm x)=\sum_{i=1}^{d}x_{i}+b^*$. We use $\mathcal{R}_{\text{lp}}(f)$ to denote the error after logit perturbation.
\begin{equation}
\begin{aligned} &\mathcal{R}_{\text{lp}}(f)\propto \Gamma\cdot\operatorname{Pr} .(\exists \| {{\tilde \delta }_{{-}}}|| \leq \epsilon
,\ \mathbb{S}(u+{{\tilde \delta }_{{-}}}) \neq-1 \mid y=-1) \\&+ \operatorname{Pr} .(\exists \| {{\tilde \delta }_{{+}}}|| \leq \epsilon\cdot\rho_+
,\ \mathbb{S}(u+{{\tilde \delta }_{{+}}}) \neq+1 \mid y=+1)
\\&=\Gamma\cdot\max _{\|{{\tilde \delta }_{{-}}}\| \leq \epsilon} \operatorname{Pr}.(\mathbb{S}(u+{{\tilde \delta }_{{-}}}) \neq-1 \mid y=-1)\\&+\max _{\|{{\tilde \delta }_{{+}}}\| \leq \epsilon\cdot\rho_+} \operatorname{Pr}.(\mathbb{S}(u+{{\tilde \delta }_{{+}}}) \neq+1 \mid y=+1)
\\&=\Gamma\cdot \operatorname{Pr}.(\mathbb{S}(u+\epsilon) \neq-1 \mid y=-1)
\\&+\operatorname{Pr}.(\mathbb{S}(u- \epsilon\cdot\rho_+)\!\neq\!+1 \!\mid\! y\!=\!+1)
\\&=\Gamma\cdot\operatorname{Pr}.\left\{\sum_{i=1}^{d}x_{i}+b+\epsilon>0 \mid y=-1\right\}
\\&+\operatorname{Pr}.\left\{\sum_{i=1}^{d}x_{i}+b-\epsilon\cdot\rho_+<0 \mid y=+1\right\}
\\&=\Gamma\cdot\operatorname{Pr}.\left\{\mathcal{N}(0,1)<-\frac{\sqrt{d}\eta}{\sigma}+\frac{b+\epsilon}{\sqrt{d} \sigma}\right\}
\\&+\operatorname{Pr}.\left\{\mathcal{N}(0,1)<-(\frac{\sqrt{d}\eta}{\sigma}+\frac{b-\epsilon\cdot\rho_+}{\sqrt{d} \sigma})\right\}.
\end{aligned}
\label{Rrob_2}
\end{equation}

The optimal $b^*$ to minimize $\mathcal{R}_{\text{lp}}(f)$ is achieved at the point
that $\frac{\partial R_{\text{lp}}(f)}{\partial b} =0$. Then we can get the optimal $b^*$:
\begin{equation}
    \begin{aligned}
    &b^* = \frac{1}{2}\epsilon(\rho-1)+\frac{d\sigma^2\text{log}\Gamma}{\epsilon-2d\eta+\epsilon\cdot\rho_+}.\label{imbal_b1}
    \end{aligned}
\end{equation}
By taking $b^*$ into $\mathcal{R}\left(f_{\text{opt}},-1\right)$ and $\mathcal{R}\left(f_{\text{opt}},+1\right)$, we can get the theorem.
\begin{equation}
    \begin{aligned}
    & \mathcal{R}\left(f_{\text{opt}},-1\right) =\operatorname{Pr}.\left\{\mathcal{N}(0,1)<-\frac{\sqrt{d}\eta}{\sigma}+\frac{b^*}{\sqrt{d} \sigma}\right\}\\&= \operatorname{Pr}.\left\{\mathcal{N}(0,1)<\frac{A}{2}+\frac{\text{log}\Gamma}{A}-\frac{\epsilon}{\sqrt{d}\sigma}\right\}, \\ & \mathcal{R}\left(f_{\text{opt}},+1\right)=\operatorname{Pr}.\left\{\mathcal{N}(0,1)<-(\frac{\sqrt{d}\eta}{\sigma}+\frac{b^*}{\sqrt{d} \sigma})\right\}\\&= \operatorname{Pr}.\left\{\mathcal{N}(0,1) <\frac{A}{2}-\frac{\text{log}\Gamma}{A}-\frac{\epsilon\cdot\rho_+}{\sqrt{d}\sigma}\right\},
    \end{aligned}\label{Imbal_Rnat1}
\end{equation}
where $A=\frac{\epsilon\cdot\rho_+-2d\eta+\epsilon}{\sqrt{d}\sigma}$.
\end{proof}
\subsection{Corollary 1}
\begin{proof}
According to Eq. (\ref{Imbal_Rnat1}), we compute the partial derivatives of $b_{\text{rob}}^*$ with respect to $\rho$ to proof the corollary.
\begin{equation}
    \frac{\partial b^*}{\partial \rho_+} =\frac{\epsilon }{2}-\frac{d\epsilon \sigma ^2\text{log}\Gamma }{(\epsilon -2d\eta +\epsilon\cdot\rho_+ )^2}. 
\end{equation}
When $\frac{\partial b^*}{\partial \rho_+}>0$, $b^*$ increases as $\rho_+$ increases. We reorganize $\frac{\partial b^*}{\partial \rho}>0$ to get the following equation.
\begin{equation}
    \begin{aligned}
    \text{log}\Gamma<\frac{(\epsilon+\epsilon\cdot\rho_+-2d\eta)^2}{2d\sigma^2}.\label{imbal_logv1}
    \end{aligned}
\end{equation}
The minimum value of the right-hand term of inequality (\ref{imbal_logv1}) is taken at $\rho_+=\frac{2d\eta-\epsilon}{\epsilon}$. But obviously, we have $\frac{2d\eta-\epsilon}{\epsilon} > \frac{\eta}{\epsilon}$. So we bring $\rho_+=\frac{\eta}{\epsilon}$ into the right-hand side of inequality (\ref{imbal_logv1}), and we get the following inequality. 
\begin{equation}
    \Gamma<e^{\frac{((2d-1)\eta-\epsilon)^2}{2d\sigma^2}}.\label{imbal_V1}
\end{equation}
When Eq. (\ref{imbal_V1}) holds, $b^*$ is a monotonically increasing function of $\rho_+$. According to Eq. (\ref{Imbal_Rnat1}), the corollary holds.
\end{proof}

\subsection{Proof for Theorem 3}
\begin{proof}
Like the proof in Theorem \ref{imbal_thm1}, we can get the following equations.
\begin{equation}
\begin{aligned} &\mathcal{R}_{\text{lp}}(f)\propto  \operatorname{Pr} .(\exists \| {{\tilde \delta }_{{+}}}|| \leq \epsilon \cdot\rho_+
, \mathbb{S}(u+{{\tilde \delta }_{{+}}}) \neq+1 \mid y=+1)\\&+\Gamma\cdot\operatorname{Pr} .(\exists \| {{\tilde \delta }_{{-}}}|| \leq \epsilon\cdot\rho_-
, \mathbb{S}(u+{{\tilde \delta }_{{-}}}) \neq-1 \mid y=-1) 
\\&=\Gamma\cdot\max _{\|{{\tilde \delta }_{{-}}}\| \leq \epsilon\cdot\rho_-} \operatorname{Pr}.(\mathbb{S}(u+{{\tilde \delta }_{{-}}}) \neq-1 \mid y=-1)\\&+\max _{\|{{\tilde \delta }_{{+}}}\| \leq \epsilon\cdot\rho_+} \operatorname{Pr}.(\mathbb{S}(u+{{\tilde \delta }_{{+}}}) \neq+1 \mid y=+1)
\\&= \Gamma\cdot\operatorname{Pr}.(\mathbb{S}(u+\epsilon\cdot\rho_-) \neq-1 \mid y=-1)
\\&+ \operatorname{Pr}.(\mathbb{S}(u- \epsilon\cdot\rho_+)\!\neq\!+1 \!\mid\! y\!=\!+1)
\\&=\Gamma\cdot\operatorname{Pr}.\left\{\sum_{i=1}^{d}x_{i}+b+\epsilon\cdot\rho_+>0 \mid y=-1\right\}
\\&+\operatorname{Pr}.\left\{\sum_{i=1}^{d}x_{i}+b-\epsilon\cdot\rho_+<0 \mid y=+1\right\}
\\&=\Gamma\cdot\operatorname{Pr}.\left\{\mathcal{N}(0,1)<\frac{1}{K}(-\frac{\sqrt{d}\eta}{\sigma}+\frac{b+\epsilon\cdot\rho_-}{\sqrt{d} \sigma})\right\}
\\&+\operatorname{Pr}.\left\{\mathcal{N}(0,1)<-(\frac{\sqrt{d}\eta}{\sigma}+\frac{b-\epsilon\cdot\rho_+}{\sqrt{d} \sigma})\right\}.
\end{aligned}
\label{Rrob}
\end{equation}
The optimal $b^*$ to minimize $\mathcal{R}_{\text{lp}}(f)$ is achieved at the point
that $\frac{\partial R_{\text{lp}}(f)}{\partial b} =0$. Then we can get the optimal $b^*$:
\begin{equation}
    \begin{aligned}
    &b^* = \frac{1}{K^2-1}(\epsilon(\rho_-+K^2\rho_+)-d\eta(K^2+1)\\&+K\sqrt{(\epsilon\rho_-+\epsilon\rho_+-2d\eta)^2+2d(K^2-1)\sigma^2\text{log}(\frac{K}{\Gamma})}).
    \end{aligned}\label{optimal_brob1}
\end{equation}

Therefore, the optimal standard error rates for the two classes can be obtained respectively.
\begin{equation}
\small
\begin{aligned} & \mathcal{R}\left(f_{\text{opt}},+1\right) =\operatorname{Pr}.\left\{\mathcal{N}(0,1)<-(\frac{\sqrt{d}\eta}{\sigma}+\frac{b^*}{\sqrt{d} \sigma})\right\}
\\&=\operatorname{Pr}.\left\{\mathcal{N}(0,1)<-K\sqrt{B^2+q(K,\Gamma)}-B-\frac{\epsilon\rho_+}{\sqrt{d}\sigma})\right\}, \\ & \mathcal{R}\left(f_{\text{opt}},-1\right) =\operatorname{Pr}.\left\{\mathcal{N}(0,1)<\frac{1}{K}(-\frac{\sqrt{d}\eta}{\sigma}+\frac{b^*}{\sqrt{d} \sigma})\right\}\\&= \operatorname{Pr}.\left\{\mathcal{N}(0,1)<KB+\sqrt{B^2+q(K,\Gamma)}-\frac{\epsilon\rho_-}{K\sqrt{d}\sigma}\right\}, \end{aligned}\label{bal-Rnat1}
\end{equation}

where $B= \frac{\epsilon\rho_-+\epsilon\rho_+-2d\eta}{\sqrt{d}\sigma(K^2-1)}$ and $q(K,\Gamma)=\frac{2\text{log}(\frac{K}{\Gamma})}{K^2-1}$.

\end{proof}

\subsection{Corollary 3}

\begin{proof}
When $\rho_-=0$ and $\rho_+=0$, we have
\begin{equation}
    \begin{aligned}
    &b^* = \frac{1}{K^2-1}(-d\eta(K^2+1)\\&+K\sqrt{4d^2\eta^2+2d(K^2-1)\sigma^2\text{log}(\frac{K}{\Gamma})}).
    \end{aligned}
\end{equation}
Let $U_+$ and $U_-$ be as follows.
\begin{equation}
    U_+ = -(\frac{\sqrt{d}\eta}{\sigma}+\frac{b^*}{\sqrt{d} \sigma});\ U_- = \frac{1}{K}(-\frac{\sqrt{d}\eta}{\sigma}+\frac{b^*}{\sqrt{d} \sigma}).\label{up+1}
\end{equation}
It is easy to verify that when $Ke^{\frac{(2d\eta-\epsilon)^2}{2dK^2\sigma^2}}<\Gamma< Ke^{\frac{2d\eta^2}{(K^2-1)\sigma^2}}$,
 $U_+>U_-$ holds. Therefore we have $\mathcal{R}\left(f_{\text{opt}},+1\right)>\mathcal{R}\left(f_{\text{opt}},-1\right)$, that is, class ``$+1$'' is harder than class ``$-1$''. 

 When $Ke^{\frac{(2d\eta-\epsilon)^2}{2dK^2\sigma^2}}<\Gamma< Ke^{\frac{2d\eta^2}{(K^2-1)\sigma^2}}$, Eq. (\ref{d+}) holds.
\begin{equation}
    \frac{\partial b^*}{\partial t}=\frac{K^2\epsilon +\frac{K^2\epsilon (\epsilon +\epsilon\rho_+-2d\eta)}{K\sqrt{(\epsilon +\epsilon\rho_+-2d\eta)^2+2d(K^2-1)\sigma ^2\text{log}(\frac{K}{\Gamma} ) } } }{K^2-1}\le 0.\label{d+}
\end{equation}
When $\frac{\partial b^*}{\partial \rho_+}<=0$, the error of class ``$+1$'' decreases and the error of class ``$-1$ increases as $\rho_+$ increases. Similarly, we can also prove other cases.

\end{proof}
\vfill

\end{document}